\documentclass[10pt,twocolumn,letterpaper]{article}

\usepackage[pagenumbers]{iccv} 

\usepackage{multirow}
\usepackage{graphicx}
\usepackage{enumitem}
\usepackage{amsmath}
\usepackage{amssymb}
\usepackage{booktabs}
\usepackage{bm}
\usepackage{multirow}
\usepackage{multicol}
\usepackage{hhline}
\usepackage{xcolor}
\usepackage{colortbl} 
\definecolor{Gray}{gray}{0.9}
\usepackage[dvipsnames]{xcolor}
\usepackage{cancel}
\usepackage{comment}
\usepackage{cuted}  
\definecolor{iccvblue}{rgb}{0.21,0.49,0.74}

\definecolor{darkgray}{gray}{0.5}
\usepackage[pagebackref,breaklinks,colorlinks,allcolors=iccvblue]{hyperref}
\usepackage[accsupp]{axessibility}
\def\paperID{12620} 
\def\confName{ICCV}
\def\confYear{2025}

\title{CLOT: Closed Loop Optimal Transport for Unsupervised Action Segmentation}

\author{Elena Bueno-Benito, Mariella Dimiccoli\\
Institut de Robòtica i Informàtica Industrial, CSIC-UPC, Barcelona, Spain\\
{\tt\small \{ebueno, mdimiccoli\}@iri.upc.edu}
}

\begin{document}

\maketitle

\begin{abstract}

Unsupervised action segmentation has recently pushed its limits with ASOT, an optimal transport (OT)-based method that simultaneously learns action representations and performs clustering using pseudo-labels. Unlike other OT-based approaches, ASOT makes no assumptions about action ordering and can decode a temporally consistent segmentation from a noisy cost matrix between video frames and action labels. However, the resulting segmentation lacks segment-level supervision, limiting the effectiveness of feedback between frames and action representations. To address this limitation, we propose Closed Loop Optimal Transport (CLOT), a novel OT-based framework with a multi-level cyclic feature learning mechanism. Leveraging its encoder-decoder architecture, CLOT learns pseudo-labels alongside frame and segment embeddings by solving two separate OT problems. It then refines both frame embeddings and pseudo-labels through cross-attention between the learned frame and segment embeddings, by integrating a third OT problem. Experimental results on four benchmark datasets demonstrate the benefits of cyclical learning for unsupervised action segmentation. \footnote{\url{https://github.com/elenabbbuenob/CLOT}}
\vspace{-3mm}

\end{abstract}    
\section{Introduction}
\label{sec:intro}
\vspace{-1mm}
Unsupervised action segmentation, the task of labeling each frame in a video or a set of videos with its action class without relying on a labeled training set, has been gaining increasing attention in recent years~\cite{Ding2021,Kumar22, Li2024, Spurio2024, Xu2024, Tran23}. 
This task is critical for numerous real-world applications, including action detection in sports events, video surveillance, and robotic manipulation~\cite{he2024, zuckerman2024}. The classical approach to unsupervised action segmentation follows a multi-step pipeline, where frame embeddings are first learned and then clustered into action segments~\cite{Kukleva2019, VidalMata2021, Li2021}.
More recently, optimal transport (OT)-based methods have emerged, jointly learning action representations and clustering frames by leveraging OT theory to generate pseudo-labels for self-supervised training.~\cite{Xu2024, Kumar22, Tran23}. 

Among OT-based methods, ASOT~\cite{Xu2024} stands out as a particularly effective approach. Unlike other techniques, it does not impose assumptions about action ordering and achieves temporally consistent segmentation by solving a Gromov-Wasserstein OT problem between video frames and action labels. However, despite its strong performance, ASOT still struggles with accurately detecting action boundaries, particularly for short-duration segments, as it implicitly biases segment length through its prior assumptions. To address this, an alternative approach, HVQ~\cite{Spurio2024}, reformulates action segmentation as a hierarchical vector quantization problem, learning a codebook to represent action clusters. While this method improves the detection of short segments, it lacks effective feedback between representation learning and clustering, typical of OT-based formulations, resulting in poorer clustering assignments.

\begin{figure}[t]
    \centering   
    \includegraphics[width=\columnwidth]{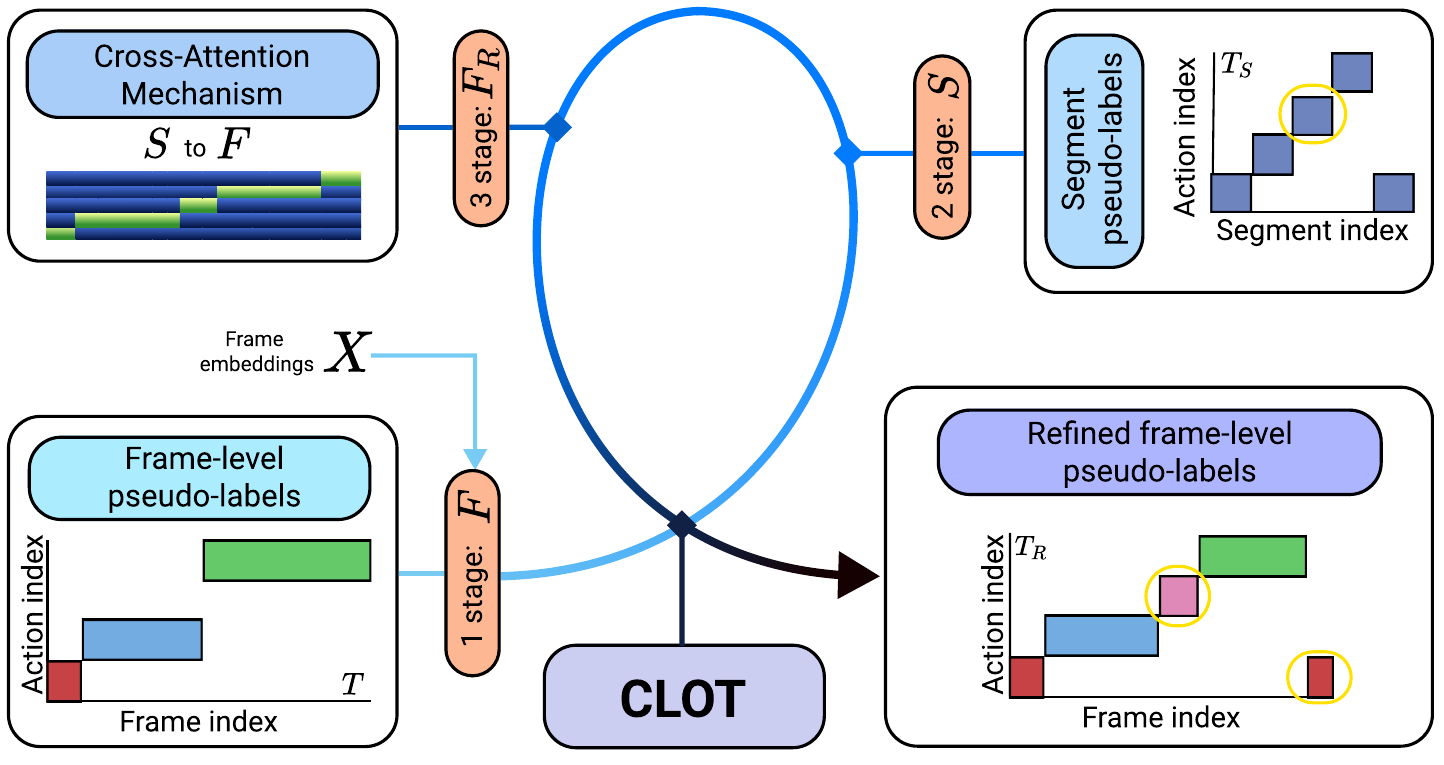}    
    \vspace{-1.2em}
    \caption{\footnotesize Given precomputed frame embeddings $X$ as input, CLOT first solves an OT problem to generate pseudo-labels $T$ alongside initial frame embeddings $F$. It then estimates segment embeddings $S$ and pseudo-labels $T_S$ by solving a second OT problem. Finally, a third OT problem is formulated to obtain refined frame embeddings $F_R$ and pseudo-labels $T_R$, by leveraging cross-attention between the learned $F$ and $S$, thereby closing the loop between frame and segment representations.
    }
    \vspace{-1.3em}
    \label{fig: CLOT}
\end{figure}

We argue that achieving high-quality, fine-grained segmentation requires \textit{explicit feedback} between frame and segment representations to ensure that learned clusters align with true segment boundaries at the video level. To this end, we propose Closed Loop Optimal Transport (CLOT), a novel OT-based framework that introduces a multi-level cyclic learning mechanism to enhance segmentation quality. 

Our approach, as illustrated in Fig.~\ref{fig: CLOT}, 
initially generates pseudo-labels and jointly learns the frame embeddings. These embeddings are then used to produce segment embeddings. Both are optimized by solving two separate OT problems. To further improve segmentation consistency, CLOT refines the pseudo-labels through cross-attention between frame and segment representations, by integrating a third OT problem.

\begin{figure*}[t]
    \centering   
    \includegraphics[width=0.90\linewidth]{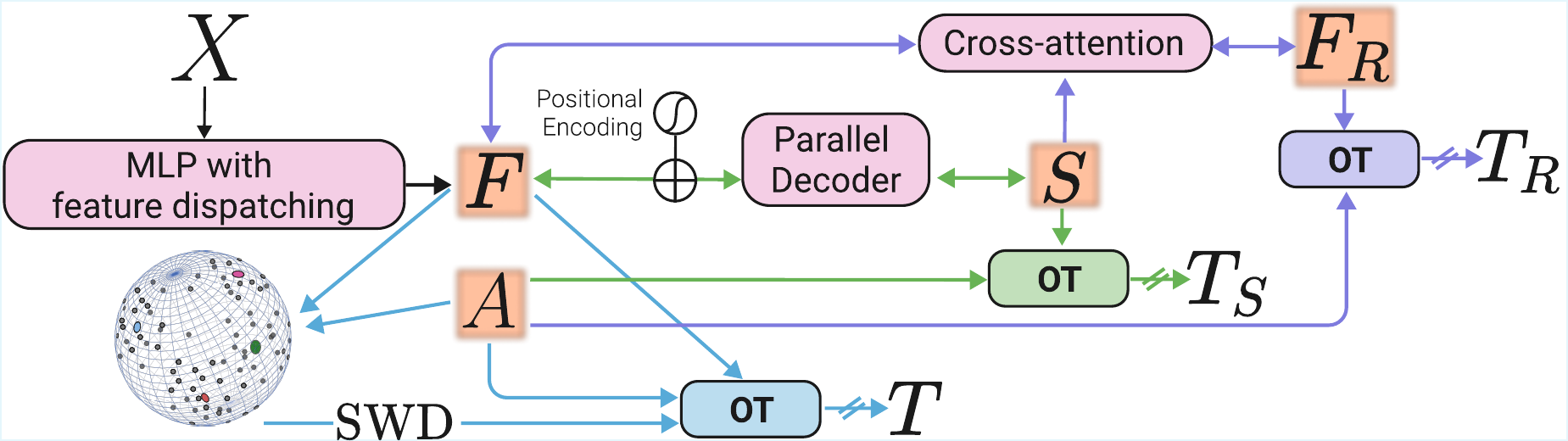}    
    \vspace{-0.7em}
    \caption{\footnotesize Diagram of the \textbf{CLOT architecture}. The only model input is the initial feature embeddings, $X$, extracted from video frames.  The model consists of a multi-layer perceptron (MLP) encoder with a feature dispatching mechanism, a parallel decoder, and a cross-attention module. The \textcolor{VioletRed}{\textbf{magenta}} boxes indicate the  \textcolor{VioletRed}{\textbf{architectural components}} of the model, whereas the \textcolor{YellowOrange}{\textbf{orange}} boxes correspond to the \textcolor{YellowOrange}{\textbf{learnable variables}}, which include the learnable action embeddings $A$ as well as the learnable frame and segment embeddings. Arrows denote the computational/gradient flow, with crossed-out arrows (\cancel{$ \rightarrow $}) indicating points where the gradient flow is stopped. Specifically, \textcolor{cyan}{\textbf{cyan}} arrows correspond to the 
    \textcolor{cyan}{\textbf{first stage}}, \textcolor{YellowGreen}{\textbf{green}} arrows to the \textcolor{YellowGreen}{\textbf{second stage}}, and \textcolor{Orchid}{\textbf{purple}} arrows to the \textcolor{Orchid}{\textbf{third stage}}. The model leverages a closed-loop optimal transport mechanism to refine frame embeddings through cross-attention with segment embeddings, ensuring improved action segmentation accuracy. The estimated OT matrices ($\mathbf{T}, \mathbf{T}_{R}, \mathbf{T}_{S}$) act as pseudo-labels during training, and are computed at different stages by using the frame/segment features ($F$, $S$, and $F_R$) together with the action embeddings $A$, which are used to define the OT cost matrices.}
    \vspace{-1.em}
    \label{fig:architecture}
\end{figure*}

Our cyclic optimization framework employs a carefully designed encoder-decoder architecture (see Fig. \ref{fig:architecture}), comprising an MLP-based encoder equipped with a feature dispatching mechanism, a parallel decoder, and a cross-attention-based mechanism.
Additionally, we incorporate the Sliced Wasserstein (SW) distance to refine the initial cost matrix, leveraging its efficiency and robustness in high-dimensional representation spaces. 
Our contributions are as follows:
\begin{enumerate}
  \item We propose a novel framework for unsupervised action segmentation that enforces segment-level consistency, achieving fine-grained segmentation and generalization across different videos.
  \item We devised a novel encoder-decoder-based architecture that closes the loop between frame and segment embeddings by leveraging their cross-attention to refine frame embeddings.
  \item We achieve very competitive performance at both video-level and activity-level over four benchmark datasets, namely Breakfast~\cite{breakfast}, YouTube INRIA~\cite{ytii}, 50salads~\cite{50salads}, and Desktop Assembly~\cite{Kumar22}.
  \item We provide extensive experiments, an ablation study, and qualitative results on four benchmark datasets, validating all components of the proposed architecture.
 
\end{enumerate}

\section{Related work}
\label{sec:SOTA}
Fully supervised action segmentation methods remain the most reliable but require costly data annotations \cite{huang2025, Bahrami2023, behrmann2022, Farha2019, Li2020, Lu_2024, Yi2021}. To improve scalability and practicality, research has increasingly shifted towards weakly-supervised \cite{Chang2021, Lu2021, Lu2022, Bin2021, Souri2022, Xu_Weak2024, Zhang2023, Bueno-Benito2024} and unsupervised approaches \cite{Bueno-Benito2023, Ding2021, Kukleva2019, dias2018learning, Kumar22, Li2021, Li2024, Sarfraz2021, Sener2018, Tran23, VidalMata2021, Xu2024, Spurio2024}, aiming to reduce reliance on labeled data while maintaining competitive segmentation performance.

\paragraph{Unsupervised activity-level action segmentation.}
Unsupervised action segmentation traditionally follows a two-step pipeline: first, learning action representations in a self-supervised manner and then clustering the learned embeddings, typically assuming a predefined number of clusters. Classical methods strongly rely on temporal regularization to model the sequential nature of activities~\cite{Kukleva2019, Sener2018}. This idea has been further refined through encoder-decoder architectures, incorporating either visual reconstruction losses (VTE)~\cite{VidalMata2021} or discriminative embedding losses (UDE)~\cite{Swetha2021} to enhance clustering performance. Other methods have framed the problem as a self-supervised learning task, where action prototypes are discovered via auxiliary classification objectives~\cite{Ding2021, Li2021}. For instance, CAD~\cite{Ding2021} introduced a framework that identifies action prototypes using an activity classification task, while ASAL~\cite{Li2021} proposed a method that distinguishes between valid and invalid action orderings based on shuffled segment predictions.
 
Recently, Optimal Transport (OT) has emerged as a powerful tool for jointly learning action representations and pseudo-labels through self-training, enabling effective feedback between representation learning and clustering while directly optimizing for action segmentation. TOT~\cite{Kumar22} introduced a temporal OT formulation that generates pseudo-labels from predicted cluster assignments. However, this approach assumes fixed action ordering across all videos and uniform label assignment, contradicting the natural variability and long-tailed distribution of action labels. UFSA~\cite{Tran23} addressed these limitations by incorporating frame- and segment-level cues from transcripts, allowing for action permutations within activities and non-uniform label assignments. However, it still requires prior knowledge of an estimated action order to infer the segmentation.

To overcome this constraint, ASOT~\cite{Xu2024} introduced an OT-based method capable of producing temporally consistent segmentations without any prior assumptions about action order. This makes it particularly suited for both pseudo-labeling and decoding. However, ASOT enforces a strong structural prior, limiting its ability to detect short-duration actions, which are critical in many real-world applications. More recently, HVQ~\cite{Spurio2024} tackled this issue by using a hierarchical vector quantization-based approach, significantly improving short-action detection. However, its learned codebook that represents action classes has limited action generalization capabilities compared to ASOT since it lacks explicit feedback between representation and clustering.



\paragraph{Unsupervised video-level action segmentation.}
Video-level action segmentation focuses on processing individual videos independently, without relying on predefined activity categories. Existing approaches can be broadly categorized into representation learning methods and clustering-based methods. Representation learning approaches aim to learn robust action features before applying a clustering algorithm. LSTM+AL~\cite{Aakur2019} predicts future frames and assigns segmentation boundaries based on prediction errors, while TSA~\cite{Bueno-Benito2023} introduced a contrastive learning framework that employs a triplet selection strategy. Clustering-based approaches directly segment videos using similarity metrics. While clustering has been underexplored in action segmentation, recent work on TW-FINCH~\cite{Sarfraz2021} incorporates temporal proximity alongside semantic similarity for improved clustering. Similarly, ABD~\cite{Zexing2022} detects action boundaries by measuring adjacent frame similarities, and OTAS~\cite{Li2024} enhances boundary detection by incorporating object-centric features.

\vspace{-1.2em}

\paragraph{Optimal Transport for Structured Prediction.}
Optimal Transport (OT) has become a key framework in machine learning for measuring distributional discrepancies, particularly in unsupervised clustering and representation learning \cite{Xu2024, Kumar22, Tran23}. Unlike traditional probability metrics, Wasserstein distance preserves the geometric structure of distributions, making it well-suited for structured prediction tasks \cite{rabin2012wasserstein, arjovsky2017, lee2018wasserstein}. However, its computational complexity limits its scalability. To address this, projection-based OT (POT) methods, such as Sliced Wasserstein (SW) distance, efficiently approximate OT by projecting high-dimensional distributions onto lower-dimensional subspaces. SW distance has been successfully applied to point-cloud processing, color transfer, Gaussian Mixture Model learning, and domain adaptation, making it a practical alternative for large-scale applications \cite{kolouri2019generalized, wu2019sliced, deshpande2019maxsliced, nguyen2023sliced, lee2019sliced}.

ASOT’s structural priors hinder short action detection and limit segmentation granularity, while its lack of feedback between frame and segment embeddings leads to suboptimal clustering. Motivated by these challenges, we propose CLOT, a novel OT-based architecture that enhances segmentation performance by better capturing the underlying video structure through a multi-level cyclic feature learning mechanism. 

\section{Closed Loop Optimal Transport}
\label{sec:method}

\paragraph{Problem formulation.}
Given a dataset $\mathcal{V} := \{V^b\}_{b=1}^{B}$ consisting of $B$ videos and, for each video $V^{b}$, an initial frame-level embeddings $X^{b}\in \mathbb{R}^{N\times D}$, where $N$ is the number of video frames and $D$ the dimension of embedding,  our goal is to learn a set of $K$ action cluster centroids, represented as $A:=[a_{1}, a_{2},\dots,a_{K}]\in \mathbb{R}^{K\times d}$, with $a_j\in\mathbb{R}^{d}$ corresponding to the centroid embedding of the $j$-th action. 

CLOT learns these centroids in an unsupervised fashion, through a three-level cyclic architecture (see Fig.~\ref{fig:architecture}). Each level solves an OT problem that learns a soft assignment: 1) between frames and action clusters in the first level, 2) between segments and action clusters in the second level, and again 3) between frames and action clusters in the third level. In the latter level, the OT problem is based on refined frame-embeddings derived from the cross-attention between frames and segment embeddings learned in the previous levels. 
The solution of each OT is strongly rooted in ASOT but differs from it in the way the initial cost matrix is computed. 
In this section, we first detail our CLOT architecture and then explain how we solve the OT problem and train the proposed network.

\subsection{CLOT architecture}

The proposed architecture, illustrated in Fig. \ref{fig:architecture}, consists of an MLP-based encoder with a feature dispatching mechanism and a parallel decoder. The encoder outputs a frame embedding $F$ by solving a temporally consistent OT problem that operates on the cost matrix (frame-to-cluster) $\mathbf{C}$. The frame embedding $F$ is then decoded into a segment embedding $S$ by the parallel decoder, which also solves a temporally consistent OT problem based on the cost matrix (segment-to-cluster) $\mathbf{C}_{S}$. A cross-attention mechanism between the encoder output $F$ and the decoder output $S$ is used to refine the frame embeddings into $F_R$, while solving a projection-based OT problem that operates on the cost matrix (frame-to-segment-to-cluster) $\mathbf{C}_{R}$. In the following, we detail each component.

\paragraph{Encoder with feature dispatching.} The encoder consists of a multi-layer perceptron network (MLP) that processes the input video frames $X \in \mathbb{R}^{N\times D}$ into frame-level embedding $F\in \mathbb{R}^{N\times d }$, where $N$ is the number of video frames and $d$ the feature dimension. The proposed method employs an adaptive process that dispatches each frame within a cluster based on similarity, resulting in a more structured and meaningful representation of the overall structure and context within the cluster, close to~\cite{liang2023clusterformer}. The feature dispatching mechanism is driven by a learnable similarity function $\phi$ parameterized by $\alpha$ and $\beta$, defined as:
\begin{equation}
    \phi(A, F)\,=\,\sigma\left ( \beta + \alpha \cdot \frac{A \cdot F}{\left\| A\right\| \left\| F\right\|}\right ),
\end{equation} 
where $\sigma$ denotes the sigmoid activation function and  $A$ represents the cluster embeddings. Each frame is assigned to a cluster using an attention-based masking approach, selecting the most relevant cluster and updating its feature representation. For each video frame $f_i \in F$, the updated feature embedding $f'_i \in F'$, with $ F'\in \mathbb{R}^{N\times d }$ is computed as:
\begin{equation}
    f'_{i}\,=\,f_{i}\,+\,\frac{1}{K}\sum_{k=0}^K (\phi(A_k, f_i)*A_{k})
\end{equation}
where $K$ is the total number of clusters and $i \in [1,...N]$. This update ensures that feature representations are dynamically adjusted based on the learned similarities, leading to more structured and meaningful frame embeddings. 

To simplify notation, the updated feature embedding $F'$ is redefined as $F$, indicating that $F$ now represents the encoder with the integrated feature-dispatching mechanism.

\paragraph{Parallel decoder.} The second stage of our framework employs a parallel decoder, adapted from \cite{Gong22,Nawhal22} for the action anticipation task. This decoder is responsible for transforming the frame-level embeddings $F$ into segment-level representations $S$ in a parallel, structured and efficient manner. Unlike autoregressive decoders like in ~\cite{behrmann2022,Girdhar21}, which predict segments sequentially and suffer from error accumulation, our model adopts a parallel decoding strategy, where all action segments are predicted simultaneously.

The decoder is based on a query-based attention mechanism inspired by object detection architectures~\cite{Carion20} and includes learnable queries $Q \in \mathbb{R}^{K'\times d_{dec}}$, where $K'$ is the number of action queries and each query serves as a segment prototype. The query-based decoder also consists of multi-head cross-attention between the encoder and decoder, along with self-attention mechanisms, to obtain the final segment embedding $S\in\mathbb{R}^{K'\times d}$ with $K'\leq K$ denoting the number of detected segments.


\paragraph{Cross-attention for refinement.} By incorporating the intrinsic information from the segment embeddings extracted for the decoder and the cluster centers, the method refines the feature embedding, enhancing the overall understanding of the video's underlying structure. 
To integrate segment-level structure into frame embeddings, we apply a cross-attention mechanism
\begin{equation} F_R = F + \text{softmax}(\frac{F S^\top}{\tau \cdot\sqrt{d}}) S, \end{equation}
where $\tau$ is a temperature parameter. This cross-attention-based alignment ensures that the feature frame representations are structured according to segment embeddings, helping to bring frame-level details to the temporal segmentation process. 

\begin{figure*}
  \centering
  \includegraphics[width=0.95\linewidth]{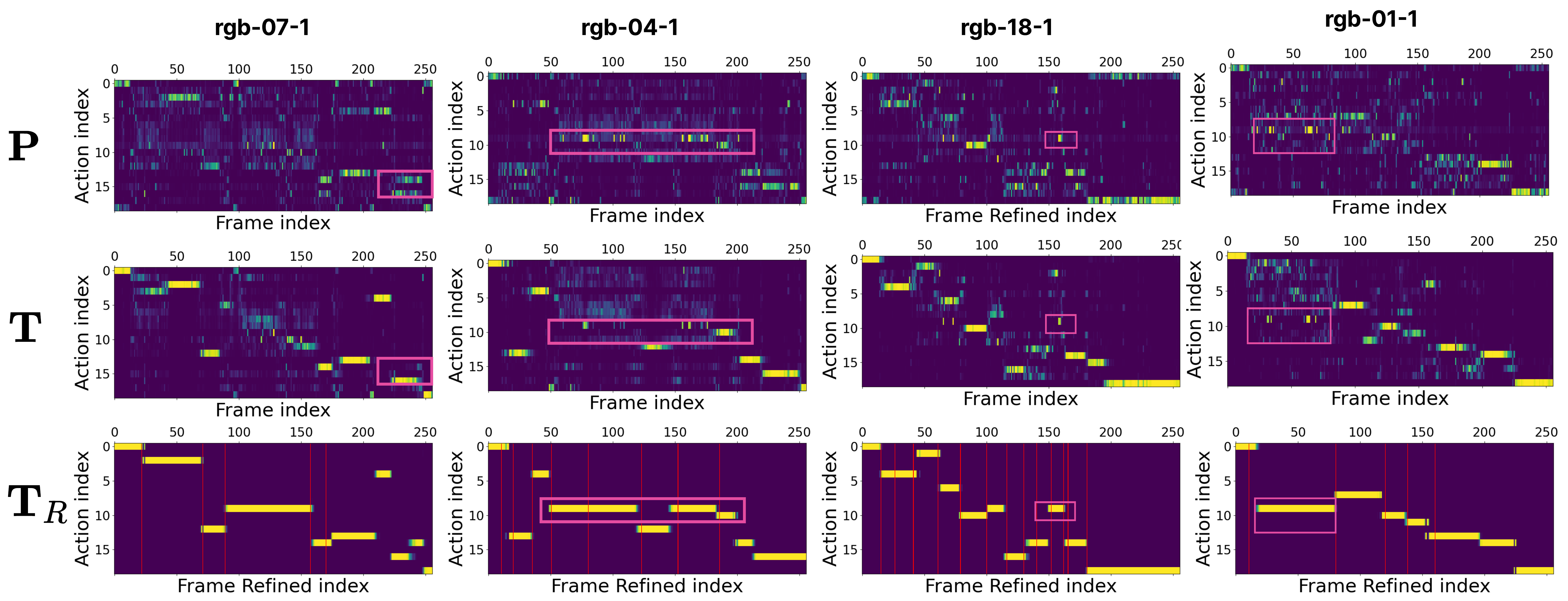}
    \caption{\textbf{Impact of cyclic multi-stage architecture.}  Each column corresponds to different videos from the 50Salads EVAL dataset (rgb07-1, rgb04-1, rgb18-1, and rgb01-1). The \textcolor{RubineRed}{\textbf{pink}} boxes highlight instances where the model ($\mathbf{T}_{R}$) accurately detects the underlying segment or refines an existing one with greater precision, compared to using only the first stage ($\mathbf{T}$) or relying solely on the predicted frame-to-action assignment probabilities ($\mathbf{P}$). The \textcolor{Red}{\textbf{red}} vertical lines denote segment boundaries. For this visualization, we use the subsampled features ($256$ per video) from the training set.}
   \label{fig:impact}
   \vspace{-4mm}
\end{figure*}

\subsection{Optimal Transport formulation}


For the sake of completeness, we recall the OT unbalanced formulation proposed in \cite{Xu2024}, which combines the classical Kantorovich Optimal Transport (KOT) with a fused Gromov-Wasserstein Optimal Transport (GWOT). 
\vspace{-0.2in}
\paragraph{KOT.} The KOT problem~\cite{thorpe2018} seeks for a transportation plan of the joint probability distribution $\mathcal{T}$ and solves for the minimum cost of coupling $\mathbf{T}^{*} \in \mathcal{T} $ between the histograms $\mu $ and $\nu$, where $\mu = \frac{1}{N}\mathbf{1}_{N} $ and $\nu = \frac{1}{K}\mathbf{1}_{K}$, being $\mathbf{1}_{N}$, and         $\mathbf{1}_{K}$ uni-dimensional vectors of ones with $N$ and $K$ dimensions, respectively. The problem is formulated as: 
\begin{equation}
\mathbf{T}^{*} = \min_{\mathbf{T}\in\mathcal{T}}\left< \mathbf{C}^{\text{k}}, \mathbf{T}\right>,
\end{equation}
\begin{equation}
 \begin{Bmatrix} \mathbf{T}\in \mathbb{R}_{+}^{N\times K}: \mathbf{T}\mathbf{1}_{K} = \mu, \mathbf{T}^{\top}\mathbf{1}_{N} =\nu
\end{Bmatrix},
\end{equation}
 where $\mathcal{T}$ represents the \textit{transportation polytope} and  $\mathbf{C}^{\text{k}}\in\mathbb{R}^{N\times K}_{+}$  is the cost matrix.
In the context of action segmentation, the variable $\mathbf{T} \in \mathbb{R}_{+}^{N\times K}$ can be interpreted as a soft assignment
between two discrete sets, corresponding to frames and action indexes.
\paragraph{GW.} The GM OT problem allows us to incorporate structural priors over the transport map, such as temporal consistency. The objective is defined as:
\begin{equation}
\mathcal{F}_{\text{GW}}(\mathbf{C}^{v},\mathbf{C}^{a},\mathbf{T}):= \sum_{i,j\in[n], j,l\in[m]} L(C^{v}_{ik},C^{a}_{jl})T_{ij} T_{kl},
\end{equation}
where $L:\mathbb{R}\times \mathbb{R}\to \mathbb{R}$ quantifies the discrepancies between the cost matrices, and $[n]$ and $[m]$ represent the set of video frames and action embedding, respectively.

The unbalanced OT formulation is as follows. 

\begin{equation}
\begin{split}
    \min_{\mathbf{T}\in\mathcal{T}_{p}} 
    \alpha \mathcal{F}_{\text{GW}}(\mathbf{C}^{v},\mathbf{C}^{a},\mathbf{T})
    +(1-\alpha) \mathcal{F}_{\text{KOT}}(\mathbf{C}^{\text{swd}},\mathbf{T})\\   
-\lambda KL(\mathbf{T}^{\top}\mathbf{1}_{n} \parallel \nu),
    \end{split}
\end{equation}
where $\alpha \in [0,1]$ balance KOT and GW terms, and $KL$ denotes the Kullback-Leibler divergence between $\mathbf{T}$ and the row-sum marginal distribution of $\mathcal{T}$ with $\lambda >0$. In particular, $KL$ penalty on the column-sum marginals of the coupling matrix $\mathcal{T}$ to $\nu$, enabling a more flexible coupling solution.


\vspace{-1mm}

\paragraph{Cost matrices.}
$\mathbf{C}:=\{ \mathbf{C}^{\text{k}},\mathbf{C}^{v},\mathbf{C}^{a}\}$ is the set of cost matrices for the KOT and GW subproblems. In our OT formulation, each of these matrices incorporates a set of submatrices:
$\{\mathbf{C}^{sw},\mathbf{C}^{v}, \mathbf{C}^{a}\}$ for frame embedding $F$ and refined frame embedding $F_{R}$ and $\{\mathbf{C}^{k},\mathbf{C}^{s}, \mathbf{C}^{a}\}$ for segment embeddings $S$. They play a crucial role in defining the optimal transport problem at each stage since they encode structural priors over the transport map desirable for the video segmentation task (i.e., temporal consistency). Except for $\mathbf{C}^{sw}$, they are defined as in \cite{Xu2024}. Specifically, the cost matrices $\mathbf{C}^{v}\in \mathbb{R}^{N\times N}$ and $\mathbf{C}^{a}\in \mathbb{R}^{K\times K}$ regulate the assignment of video frames and action categories, respectively by penalizing assignments of adjacent frames $(|i - k|\leqq Nr, i\neq k)$ to different action embedding $(j\neq l)$ for two differents assignments $T_{ij}$ and $T_{kl}$. However, no penalty is applied to assignments outside temporal radius $Nr$ or adjacent frames mapped to the same action $(j = l)$. The last cost matrix $\mathbf{C}^{k}$ is the visual component and is defined as $\mathbf{C}^{k}_{i,j}=  \frac{x_{i}^{\top}\cdot a_j}{\left\| x_i\right\| \left\| a_j\right\|} - \rho\cdot\mathbf{Z}_{ij} $ and $\mathbf{Z}$ is the temporal prior used in \cite{Kumar22, Xu2024} defined as $\mathbf{Z}_{ij} = |1/N- j/K|$ for $\rho\geq 0$.

For $\mathbf{C}^{sw}$, we introduce the
Sliced-Wasserstein (SW) distance as a complementary metric to the cosine distance, which has been successfully applied to many other structured prediction problems  ~\cite{kolouri2019generalized, wu2019sliced, deshpande2019maxsliced, nguyen2023sliced, lee2019sliced}.
The p-Wasserstein distance between two probability measures $\mu$ and $\nu$ in the probability space is defined as: 
\begin{equation}
W_p(\mu, \nu) = \left( \inf_{\mathcal{T} \in \Pi(\mu, \nu)} \int_{\mathbb{R}^d \times \mathbb{R}^d} d^p(x,y) d\mathcal{T}(x,y) \right)^{\frac{1}{p}},
\end{equation}
where  $\Pi(\mu,\nu)$ denotes the set of all couplings of $\mu$ and $\nu$, i.e., joint distributions with marginals $\mu$ and $\nu$. 
For $p\geq1$, we estimate the discrete SWD distance empirically by discretizing the integral over the unit sphere $\mathcal{S}^{d-1}$ in $\mathbb{R}^{d}$ using a finite set of random projections $\{\theta_{i}\}_{i=1}^{M}$ between $x_i\in X$ and $a_{j}\in A$, leading to:
\begin{equation}
\text{SWD}_p(x_i, a_j) = \left(  \frac{1}{M} \sum_{m=1}^{M} d( R_{\theta_m \#} x_i, R_{\theta_m \#} a_j ) \right)^{\frac{1}{p}},
\end{equation}
for $M$ randomly sampled $\theta$, $d$ quadratic loss and $\forall (i, j)\in N\times K$.
In our method, we use $p=1$, which means that it is equivalent to solving several one-dimensional optimal transport problems with closed-form solutions~\cite{rabin2012wasserstein}. 

Specifically, the cost matrix $\mathbf{C}^{\text{sw}}\in\mathbb{R}^{N\times K}_{+}$ captures the visual similarity between frames and actions, defined as
 \begin{equation}
     \mathbf{C}_{ij}^{\text{sw}} = 1 + \text{SWD}(x_i,a_j) - \mathbf{C}^{k}_{i,j}
 \end{equation}

 \subsection{CLOT learning pipeline}
 
We learn the parameters of our architecture in an unsupervised manner. The training objective minimizes a cross-entropy loss between predicted action probabilities $P$ and their corresponding soft pseudo-labels $T$ derived by solving an OT problem. See Fig.~\ref{fig:impact} to visualize the impact of our cyclic multi-stage training. Specifically, we integrate three key components: frame embeddings ($\mathbf{F}$) generated by the MLP, segment embeddings generated by the decoder ($\mathbf{S}$), and refined frame representations ($\mathbf{F}_{R}$ ) obtained by using the cross-attention between $\mathbf{F}$ and $\mathbf{S}$. Each of these components is associated with soft pseudo-label $\mathbf{T}\in \mathbb{R}^{K\times N}$, $\mathbf{T}_{S}\in \mathbb{R}^{K\times K'}$ and $\mathbf{T}_{R}\in \mathbb{R}^{K\times N}$, respectively. 
For a given video $b$, we define the predicted frame-to-action assignment probabilities as
\begin{equation}
    P^b_{ij} = \text{softmax}(\mathbf{F}^b \mathbf{A}^\top)_{ij} / \tau,
\end{equation}
where $\tau >0$ is a temperature scaling parameter. The frame-to-action learning loss is then formulated as
\begin{equation}
\mathcal {L}(\mathbf{T}, \mathbf{P}) = -\frac {1}{B}\sum _{b=1}^B \sum_{i=1}^N\sum_{j=1}^K T^b_{ij} \log P^b_{ij},
\end{equation}
Similarly, we compute the segment-to-action assignment probabilities $\mathbf{P}_{S}$ and the frame-to-segment-to-action assignment probabilities $\mathbf{P}_{R}$. Their corresponding loss functions are defined analogously. The final training objective is a sum of these three losses
\begin{equation}
\mathcal {L}_{\text {train}} =  \mathcal {L}(\mathbf{T}, \mathbf{P})+\mathcal {L}(\mathbf{T}_{S}, \mathbf{P}_{S})+\mathcal {L}(\mathbf{T}_{R}, \mathbf{P}_{R})
\end{equation}

\section{Experimental Results}
\label{sec:results}

\begin{table*}[ht!]
    \centering
    \setlength\tabcolsep{3pt}
    \resizebox{6.0in}{!}{
    \begin{tabular}{@{}lcccccccccccccccc@{}}
    \toprule
     \multirow{2}{*}{Methods} & \multicolumn{3}{c}{Breakfast} &  \multicolumn{3}{c}{YTI} &\multicolumn{3}{c}{50Salads (Mid)} & \multicolumn{3}{c}{50Salads (Eval)} & \multicolumn{3}{c}{DA} \\
    \cmidrule(lr){2-4} \cmidrule(lr){5-7} \cmidrule(lr){8-10} \cmidrule(lr){11-13} \cmidrule(lr){14-16}
    & MoF & F1 & mIoU & MoF & F1 & mIoU & MoF & F1 & mIoU & MoF & F1 & mIoU & MoF & F1 & mIoU  \\
    \midrule
     CTE~\cite{Kukleva2019} &  41.8 & 26.4 & - & 39.0 & 28.3  & - &30.2   & -  & -  & 35.5  & - & - &47.6 & 44.9& -\\
   VTE~\cite{VidalMata2021} & 48.1  & - & - & - & 29.9  &-  & 24.2  & -  & -  & 30.6  & - & - &-  &-  &- \\
    UDE~\cite{Swetha2021} & 47.4  & 31.9 & - & 43.8 & 29.6  & - & -  &  - & - &  42.2 & 34.4  & - & - & - & - \\

    ASAL~\cite{Li2021} & 52.5  & 37.9 & - & 44.9 & 32.1  & - & 34.4  & -  & -  & 39.2  &-  &-  &-  &-  &- \\
    
    TOT~\cite{Kumar22}  &47.5 & 31.0 & - &40.6 & 30.0 & - &31.8 & - & - &47.4 & 42.8 & - &56.3 & 51.7 & -   \\
    TOT+~\cite{Kumar22} &39.0 & 30.3 & - &45.3 & 32.9 & - &34.3 & - & -& 44.5 & 48.2 & - &58.1 & 53.4 & - \\
    UFSA~\cite{Tran23} &52.1 & 38.0 & - & 49.6 & 32.4 & - & 36.7 & 30.4 & - & 55.8 & 50.3 & -& 65.4 & 63.0 & - \\
     ASOT~\cite{Xu2024} & \underline{ 56.1} & 38.3 & \textbf{18.6} & \underline{52.9} & \underline{35.1} & \textbf{24.7} & \underline{46.2} & \underline{37.4} & \underline{24.9} & \underline{59.3} & \underline{53.6} & \underline{30.1} & \textbf{70.4} & \underline{68.0} & \underline{45.9}\\
    HVQ~\cite{Spurio2024} & 54.4  & \underline{39.7} & - & 50.3 & \underline{35.1}  & - & -  & -  & -  &-   & - & - & - & - & - \\

     \rowcolor{Gray} CLOT (Ours) & \textbf{60.1}   & \textbf{40.1} &  \underline{18.5}& \textbf{ 54.4} & \textbf{36.7}  & \underline{23.4} & \textbf{50.6}  & \textbf{46.6} & \textbf{31.4} &   \textbf{59.4} & \textbf{63.2} & \textbf{38.8}   & \underline{68.8} &\textbf{ 72.6}  & \textbf{48.1} \\
    
    \bottomrule
    \end{tabular}}
    \caption{Comparisons of action segmentation performance obtained by applying the \textcolor{blue}{Hungarian matching} at the \textcolor{blue}{activity-level} on the Breakfast~\cite{breakfast}, Youtube Instr. ~\cite{ytii},  50Salads ~\cite{50salads} and Desktop Assembly \cite{Kumar22} benchmarks. The highest accuracy is indicated in \textbf{bold}, and the second highest is \underline{underlined}. }
    \label{tab:sota_activity}
    \vspace{-4mm}

\end{table*}

\begin{table*}[ht!]
    \centering
    \setlength\tabcolsep{3pt}
    \resizebox{6.0in}{!}{
    \begin{tabular}{@{}lccccccccccccccc@{}}
    \toprule
    \multirow{2}{*}{Methods} & \multicolumn{3}{c}{Breakfast} &  \multicolumn{3}{c}{YTI} &\multicolumn{3}{c}{50Salads (Mid)} & \multicolumn{3}{c}{50Salads (Eval)} & \multicolumn{3}{c}{DA} \\
    \cmidrule(lr){2-4} \cmidrule(lr){5-7} \cmidrule(lr){8-10} \cmidrule(lr){11-13} \cmidrule(lr){14-16}
    & MoF & F1 & mIoU & MoF & F1 & mIoU & MoF & F1 & mIoU & MoF & F1 & mIoU & MoF & F1 & mIoU  \\
    \midrule
   
    TWF*~\cite{Sarfraz2021} & 62.7  & 49.8  & 42.3  & 56.7  & 48.2  & - & 66.8  & \underline{56.4}  & \textbf{48.7}  &\underline{71.7} &-  & -  &73.3  & 67.7  & \textbf{57.7} \\
    ABD*~\cite{Zexing2022} & 64.0 & 52.3 & - & 67.2 & 49.2 & - & \underline{71.8} & - & - & 71.2 & - & - & - & - & - \\
    OTAS*~\cite{Li2024} &\textbf{67.9} & - & - & 65.7 & - & - & \textbf{72.4} & - & - & \textbf{73.5} & - & - & - & - & - \\
    TSA*~\cite{Bueno-Benito2023} \small(kmeans) & 63.7 & \textbf{58.0} & \textbf{53.3} & 59.7& 55.3  & - &-&-&-&-&-&-&-&-&-\\
    
    ASOT~\cite{Xu2024} & 63.3 & 53.5 & 35.9 & \textbf{71.2} & \textbf{63.3} & 47.8 & 64.3 & 51.1 & 33.4 & 64.5 & \underline{58.9} & \underline{33.0} & \underline{73.4} & \underline{68.0} & 47.6\\

     \rowcolor{Gray} CLOT (Ours) & \underline{66.3} & \underline{55.9} & \underline{37.1}& \underline{69.3} & \underline{60.8} & \textbf{48.2} &  69.4 & \textbf{63.8}  & \underline{45.0} & 64.6 & \textbf{69.7} & \textbf{42.5} &  \textbf{73.5}& \textbf{75.2}  & \underline{52.4}  \\
    
    \bottomrule     

    \end{tabular}}
    \caption{Comparisons on the Breakfast~\cite{breakfast}, Youtbe Instr.~\cite{ytii},  50Salads~\cite{50salads} and Desktop Assembly \cite{Kumar22} datasets computed by applying \textcolor{blue}{Hungarian matching} per \textcolor{blue}{video}. * denotes whether the method has a training stage on target videos. The highest accuracy is indicated in \textbf{bold}, and the second highest is \underline{underlined}. }
    \label{tab:sota_video}
    \vspace{-4mm}

\end{table*}
\begin{table*}[ht!]
    \centering
    \setlength\tabcolsep{3pt}
    \resizebox{6.0in}{!}{
    \begin{tabular}{@{}lcccccccccccccccc@{}}
    \toprule
     \multirow{2}{*}{} & \multicolumn{3}{c}{Breakfast} &  \multicolumn{3}{c}{YTI} &\multicolumn{3}{c}{50Salads (Mid)} & \multicolumn{3}{c}{50Salads (Eval)} & \multicolumn{3}{c}{DA} \\
    \cmidrule(lr){2-4} \cmidrule(lr){5-7} \cmidrule(lr){8-10} \cmidrule(lr){11-13} \cmidrule(lr){14-16} 
     & MoF & F1 & mIoU & MoF & F1 & mIoU & MoF & F1 & mIoU & MoF & F1 & mIoU & MoF & F1 & mIoU  \\
    \midrule
    \rowcolor{Gray} CLOT  & 60.1  & 40.1 &  18.5 & 54.4& 36.7 & 23.4 & 50.6  & 46.6 & 31.4 & 59.4 & 63.2 & 38.8& 68.8 &72.6& 48.1 \\
    w/o SWD  & 59.8  & 39.8 & 18.0 & 53.8 &34.8 &20.7  & 47.9 & 38.5 & 25.0& 58.1 & 52.7 & 25.4 & 58.5& 62.3&38.5\\
    w/o FD  & 51.9 & 35.9 & 16.2 & 51.9& 34.6& 22.2& 45.0 & 31.2 &24.9 & 51.7 &51.9  & 26.2& 63.6& 68.3& 44.3\\
    w/o Decoder  & 58.0  & 38.9 & 18.2 & 51.5 & 35.8& 23.3  & 44.9 & 37.5 & 26.0& 57.7 & 52.7 &26.3 &66.2 &68.4 &46.5\\
    w/o Refinement  &  59.7 & 39.6 & 18.3 & 51.9& 35.3& 20.9&  47.1 & 37.8 & 25.3& 58.2 & 52.6 & 24.5&68.2 &70.6 &47.6\\


    \bottomrule     

    \end{tabular}}
    \caption{Ablation Study on the four datasets: Breakfast~\cite{breakfast}, YTI~\cite{ytii}, 50Salads~\cite{50salads} and Desktop Assembly~\cite{Kumar22}}
    \label{tab:ablation_study}, 
    \vspace{-7mm}

\end{table*}
\subsection{Experimental setting}
\paragraph{Datasets and Features.} We evaluate our approach using four widely used video datasets, each with pre-extracted feature representations for fair comparison with prior work~\cite{Tran23, Kukleva2019, Kumar22, Spurio2024, Xu2024, Ding2021}:

\begin{itemize}
    \item \textbf{Breakfast (BF)}~\cite{breakfast} comprises approximately $1,700$ videos depicting individuals preparing various breakfast items. Each video belongs to one of $10$ activity categories, covering $48$ distinct actions such as \textit{cracking an egg} or \textit{pouring flour}. The videos range in duration from $30$ seconds to several minutes. We use Fisher vector features extracted from IDT~\cite{IDT}.
    
    \item \textbf{YouTube Instructions (YTI)}~\cite{ytii} consists of $150$ instructional videos spanning $5$ activity categories. These videos, averaging around $2$ minutes in length, primarily focus on tutorials, with a significant portion of frames ($\approx 75\%$) containing background content. We use a concatenation of Histogram of Optical Flow (HOF) descriptors and features extracted from the VGG16-conv5 layer~\cite{VGG}.
    
    \item \textbf{50 Salads (FS)}~\cite{50salads} consists of $50$ videos, totaling $4.5$ hours, in which actors perform cooking activities. The dataset provides two levels of granularity: Mid ($19$ action classes) and Eval ($12$ action classes), where some Mid-level actions are aggregated into single actions in the Eval level. We extract Fisher vector features from IDT~\cite{IDT}.
    
    \item \textbf{Desktop Assembly (DA)}~\cite{Kumar22} includes $76$ videos, each approximately $1.5$ minutes long, depicting actors assembling items in a predefined order. The task involves $22$ sequentially performed actions. We use the features provided by~\cite{Tran23}.
\end{itemize}
\paragraph{Metrics.} 
We follow established evaluation protocols for unsupervised action segmentation, consistent with prior studies~\cite{ding2023survey}. To align learned action clusters with ground truth labels, we apply Hungarian Matching at the video level~\cite{Sarfraz2021, Zexing2022, Bueno-Benito2023}, where matching is performed within each video individually, and at the activity level~\cite{Tran23, Kukleva2019, Kumar22, VidalMata2021, Spurio2024, Xu2024}, where matching is performed across all videos of the same activity category.%

To assess action segmentation quality, we employ the following metrics: (i) \textbf{Mean over Frames (MoF)}, which represents the percentage of correctly predicted frames and is sensitive to class imbalance, favoring dominant classes; (ii) \textbf{F1 Score}, the harmonic mean of precision and recall, as proposed in~\cite{Kukleva2019}, which complements MoF by treating large segments as a whole and mitigating the effects of class imbalance; and (iii) \textbf{Mean Intersection over Union (mIoU)}, which computes the average intersection-over-union across all classes, explicitly addressing the impact of class imbalance.


\vspace{-4mm}

\paragraph{Implementation details}
The encoder employs an MLP with a single hidden layer, while the decoder follows a parallel architecture. Dropout rates, projection dimensions, the number of predicted segments $K'$ per video and other dataset-specific hyperparameters are adjusted for each dataset, with full details provided in the supplementary material. Optimization is performed using the Adam optimizer with a learning rate of $10^{-3}$ and weight decay of $10^{-4}$. Action embeddings are initialized through k-means clustering. Training is conducted with $256$ frames per video, sampled from uniformly distributed intervals. The number of action clusters $K$ is defined based on the ground truth action count for each dataset, ensuring consistency with prior work~\cite{Tran23, Kukleva2019, Kumar22, VidalMata2021, Spurio2024, Xu2024}.
\subsection{State-of-the-Art Comparisons} 
\begin{figure}[t]
    \centering
    \begin{subfigure}[t]{\linewidth}
    \vspace{-1mm}
    \includegraphics[width=\columnwidth]{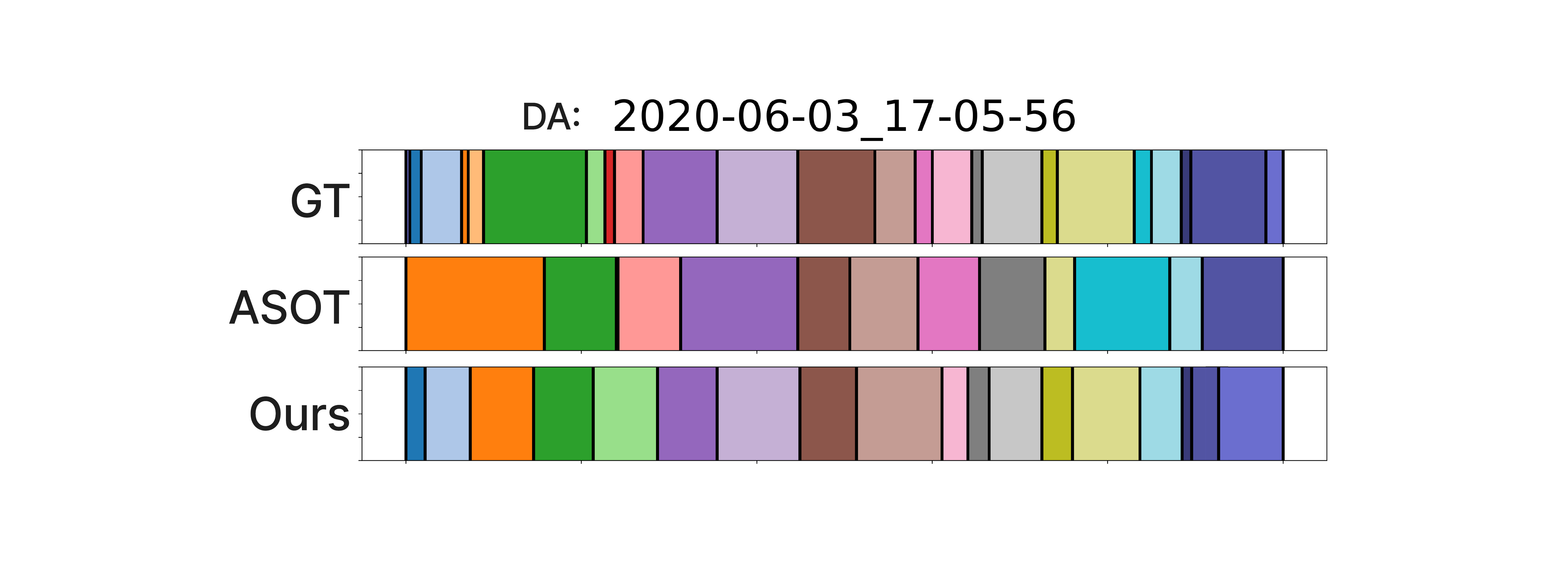}
    \end{subfigure}
    \begin{subfigure}[t]{\linewidth}
    \vspace{-7mm}
    \includegraphics[width=\columnwidth]{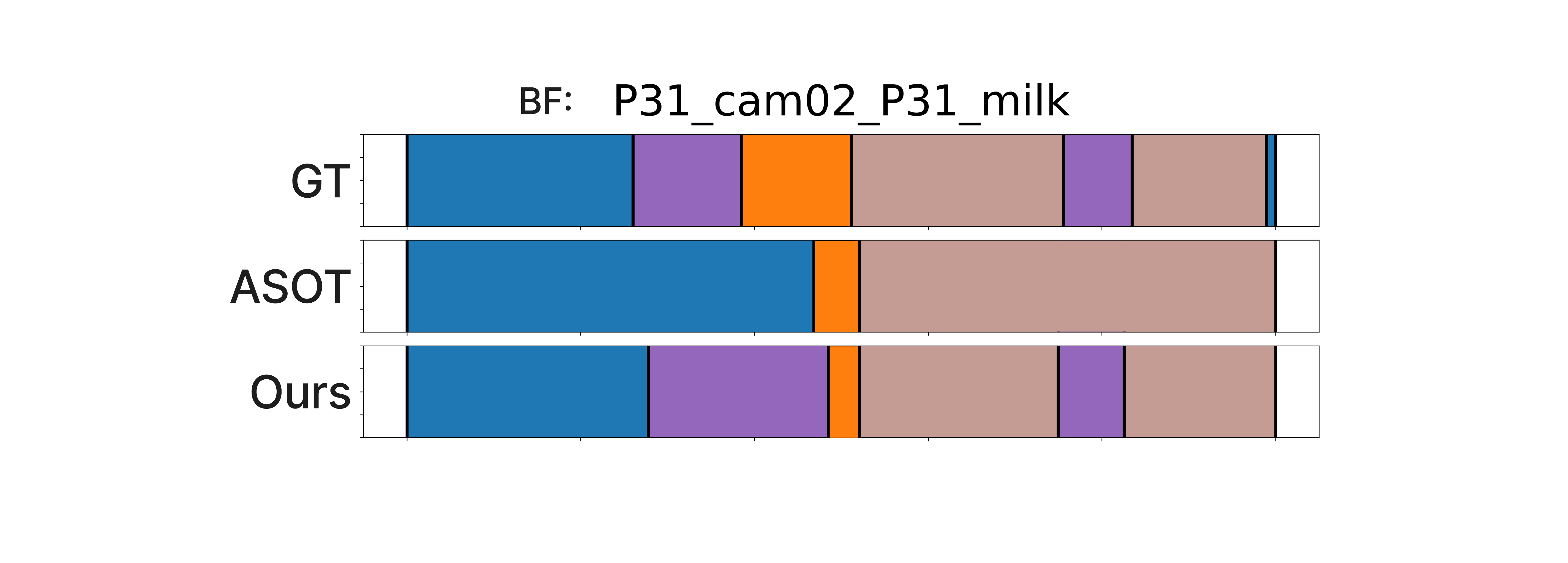}
    
    \end{subfigure}
        \begin{subfigure}[t]{\linewidth}
    \vspace{-8mm}
    \includegraphics[width=\columnwidth]{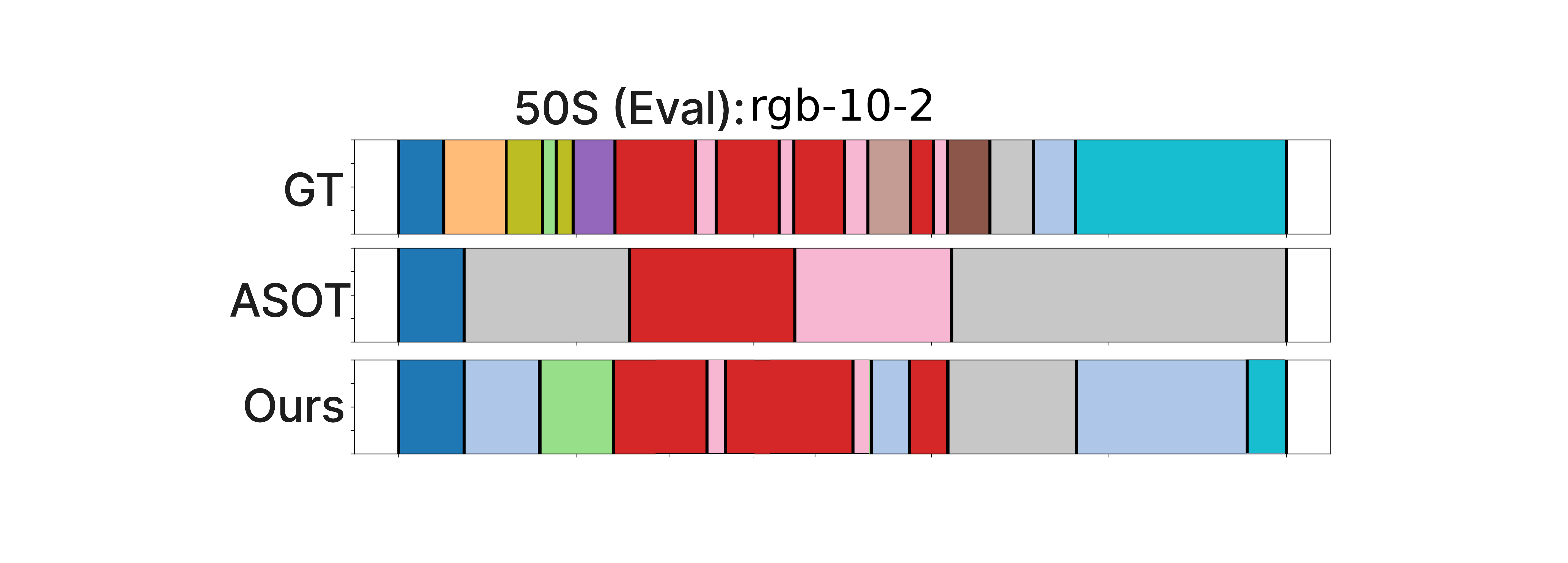}
    \vspace{-2mm}
    \end{subfigure}
    \begin{subfigure}[t]{\linewidth}
    \vspace{-10mm}
    \includegraphics[width=\columnwidth]{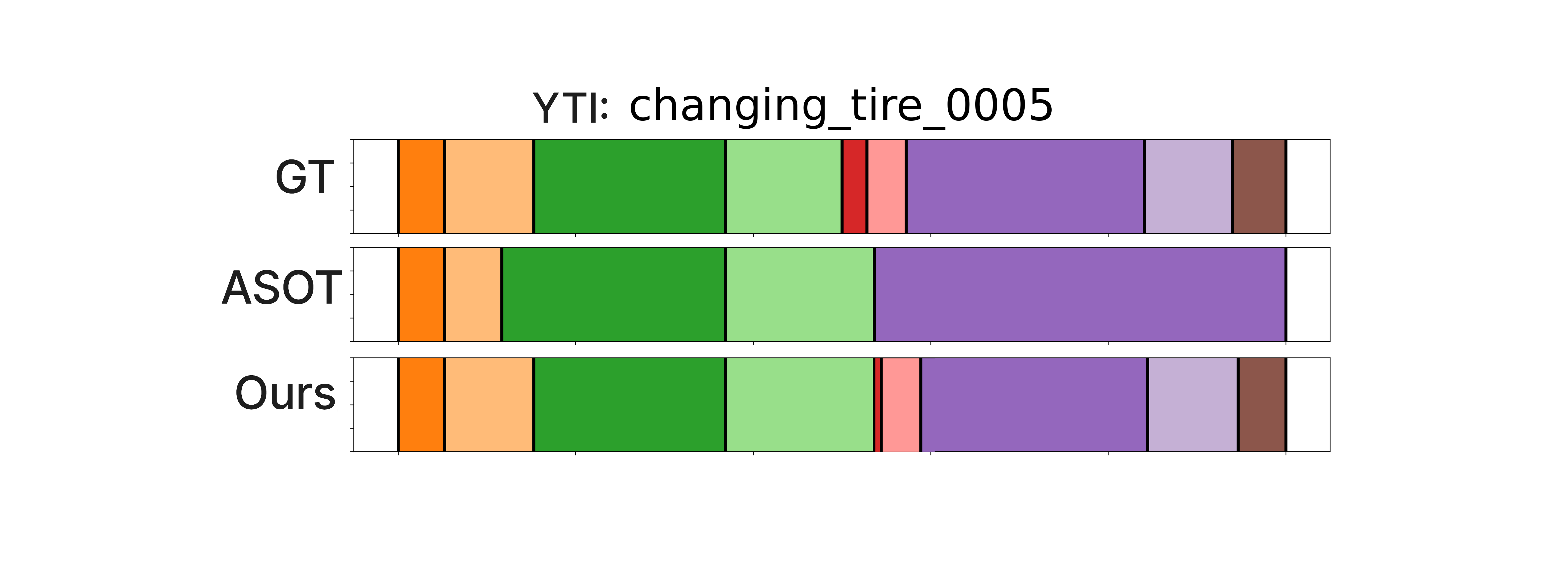}
    \vspace*{-2mm}
    \end{subfigure}
     
    \par \vspace*{-5mm}
    
\caption{\textbf{Qualitative results.} We display the ground-truth (GT), the results of CLOT (Ours), and ASOT~\cite{Xu2024}. Results from different datasets and activities are displayed for comparison.}

\vspace*{-4mm}
 \label{fig:qualitative}
\end{figure}

Traditionally, unsupervised action segmentation at the video level and the activity level have been treated as separate challenges. Video-level segmentation is generally considered the less complex of the two, as it does not require generalizing action representations across multiple videos. In contrast, activity-level segmentation demands the ability to maintain accurate action boundary detection while ensuring generalization across different instances of the same activity. This trade-off has historically limited performance improvements across both settings.
CLOT establishes a new state-of-the-art in both video-level and activity-level segmentation. It consistently outperforms previous approaches on almost all benchmark datasets. Below, we analyze our results in detail, comparing them to existing methods and interpreting the architectural contributions that drive our superior performance. Fig~\ref{fig:qualitative} contains qualitative examples where actions are seen in different orders across videos within a dataset. More results are in the supp. mat.

\paragraph{Activity-level.} CLOT surpasses all existing methods, achieving a significant performance gain. Specifically, our approach improves over the previous best-performing method by an average of $+2.54\%$ in MoF, $+5.08\%$ in F1 Score, and $+3.2\%$ in mIoU across five benchmark datasets. 

Our F1 results consistently outperform the state of the art, with a particularly large margin in 50Salads and DA. This suggests that our action embeddings are more descriptive than the cluster embeddings from ASOT~\cite{Xu2024} or the vector quantization from HVQ~\cite{Spurio2024}. However, our mIoU scores are marginally lower than those of ASOT on the BF and YTI datasets. The mIoU metric assigns equal weight to all classes, meaning that infrequent actions disproportionately affect the final score. If certain actions are never detected due to their rare occurrence, mIoU is more affected than MoF or F1 Score. This explains why our F1 scores remain high even in datasets where mIoU is slightly lower. Nonetheless, in the remaining datasets, we demonstrate a substantial improvement in mIoU ($\sim 5.8\%$ on average). 

CLOT lies in its parallel decoder architecture, which ensures that segmentation errors do not propagate across an entire sequence. Unlike autoregressive models, where errors accumulate over time, our parallel decoder prevents error propagation, ensuring that a single misclassified segment does not negatively impact the entire sequence. This allows the model to effectively avoid segmentation drift while maintaining high-quality action boundaries (comparisons in supp. mat).
\vspace*{-2mm}

\paragraph{Video-level}
CLOT and ASOT~\cite{Xu2024} learns a single model across all videos of an activity. In contrast, * methods are trained and optimized independently per video and are tailored to each test instance, often achieving higher performance by design, without generalization across videos. CLOT outperforms ASOT~\cite{Xu2024} across all datasets except for YTI, while often achieving F1 and mIoU scores comparable to methods specifically designed for video-level segmentation. 
This improvement suggests that CLOT dynamically refines action representations, allowing it to adapt more effectively to varying segment structures, which is particularly important at the video-level.

\subsection{Ablation Study}
To assess the impact of each component of our proposed framework, we conducted an ablation study on four benchmark datasets by systematically removing key elements from our architecture and measuring the resulting performance degradation. Further ablation experiments and sensitivity analysis are provided in the supp. mat. 
The results are reported in Tab~\ref{tab:ablation_study}. Our full model, CLOT, achieves the highest performance across all datasets. 

Removing SWD leads to a decline in performance across all datasets, particularly in terms of mIOU. This suggests that SWD plays a critical role in enhancing the inicial cost matrix. The performance drop is evident in the 50Salads and DA datasets. By projecting data onto one-dimensional subspaces, SWD helps the visual information to reduce the error in videos where actions have a strong bias between actions. 

Removing the feature dispatching (FD) mechanism results in a noticeable decrease in MoF and F1. This suggests to us FD prioritises informative frames and filters out irrelevant and noise frames (no visual information or background).
Without the Decoder, Mof and mIoU drop significantly, particularly in 50S and YTI, suggesting that segment-level embeddings are essential for short-duration actions. The decoder improves structured segmentation and enables a more global representation of actions, complementing frame-level embeddings.

Eliminating the refinement stage results in consistent performance degradation, particularly in F1 and mIOU. This suggests the importance of incorporating segment-level structure to refine frame embeddings and action boundaries. 

The results demonstrate that each of the introduced components adds value to the final segmentation results, even if the magnitude of the impact may vary significantly depending on the dataset. This study confirms that the integration of these components mitigates reliance on strong structural priors, filters noisy frames, and enables a more comprehensive action representation.


\section{Conclusion}
\vspace*{-1.2mm}
\label{sec:conclusion}
We presented CLOT, a novel Optimal Transport-based framework for unsupervised action segmentation that introduces a multi-level cyclic learning mechanism to refine frame and segment embeddings. Unlike previous methods, CLOT closes the loop between representation learning and clustering, ensuring a structured and adaptive segmentation process. 
The key architectural components of CLOT are the enhancement of the initial cost matrix through the Sliced Wasserstein Distance, which captures finer inter-frame relationships, a feature dispatching mechanism that filters out noisy or background frames, a parallel decoder that provides segment-level estimations, and a cross-attention mechanism that allows refining frame-level embedding. Altogether, these components ensure high-quality segmentation accuracy across dataset variability. Our experiments on four benchmark datasets confirm the effectiveness of CLOT, consistently surpassing existing methods in both video-level and activity-level segmentation.

\noindent \textbf{Limitations.} While CLOT achieves state-of-the-art performance, action mislabeling remains the main open challenge, highlighting the need for improvements in pseudo-label refinement and segment alignment. We believe CLOT represents a significant step forward in OT-based action segmentation, paving the way for advances in self-supervised learning and structured prediction tasks.
 
\vspace{-0.50em}
\section*{Acknowledgements}
\vspace{-0.50em}
This work was supported by grant \small{PRE2020-094714}, the project \small{PID2019-110977GA-I00} and the project \small{PID2023-151351NB-I00}, funded by Ministerio de Ciencia e Innovación (MCIN)/ Agencia Estatal de Investigación (AEI) /10.13039/501100011033, by European Social Fund (ESF) Investing in your future and by \textit{ERDF, UE}. It was also supported by an \textit{Alexander von Humboldt (AvH)} fellowship for experienced researchers funded by the \textit{AvH Foundation}.

{
    \small
    \bibliographystyle{ieeenat_fullname}
    \bibliography{main}
}
\clearpage
\appendix
\begin{strip}
    \centering
    {\LARGE \textbf{Supplementary Material for}}\\[0.5em]
    {\Large \textbf{CLOT: Closed Loop Optimal Transport for Unsupervised Action Segmentation}}\\[2em]
    
    \textbf{Elena Bueno-Benito, Mariella Dimiccoli}\\[0.5em]
    Institut de Robòtica i Informàtica Industrial, CSIC-UPC, Barcelona, Spain\\
    \texttt{\{ebueno, mdimiccoli\}@iri.upc.edu}
\end{strip}

This supplementary material provides extended analyses and implementation details to support our proposed CLOT paper. It includes sensitivity studies on key parameters, complexity comparisons with ASOT, and additional ablations to isolate the effect of each component. We also provide full hyperparameter settings and additional qualitative results across all datasets to illustrate the advantages of our approach further.

\section{Sensitivity Analysis}

\begin{figure*}[h]
    \centering
    \hspace*{-2em}
    \begin{subfigure}[t]{0.7\linewidth}
        \includegraphics[width=1.05\linewidth]{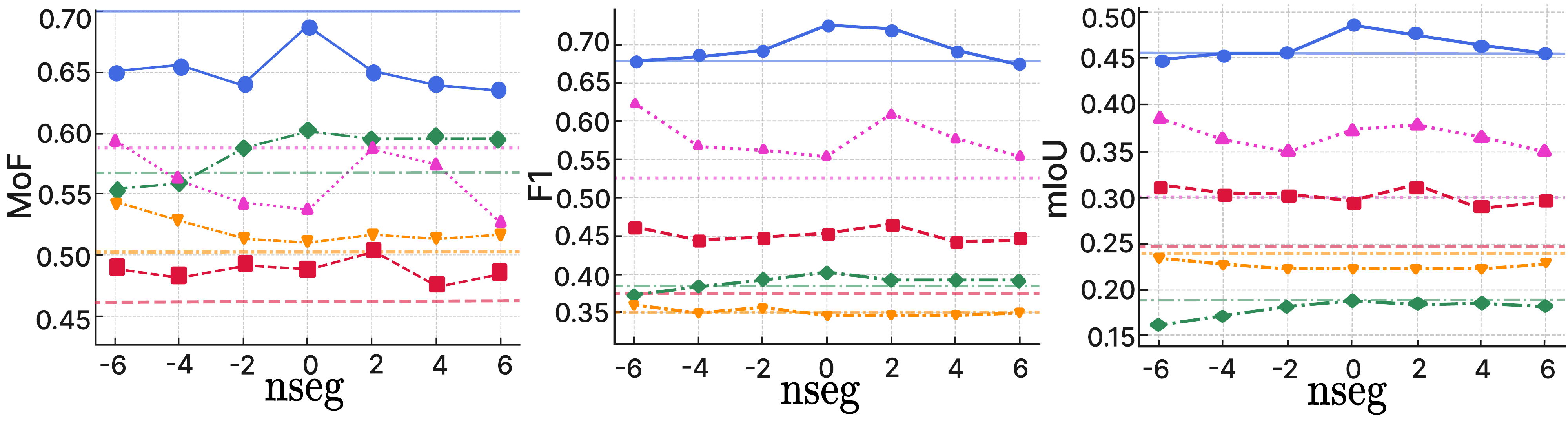}
    \end{subfigure} 
    \hspace*{1em}
    \begin{subfigure}[t]{0.3\linewidth}
        \includegraphics[width=0.9\linewidth]{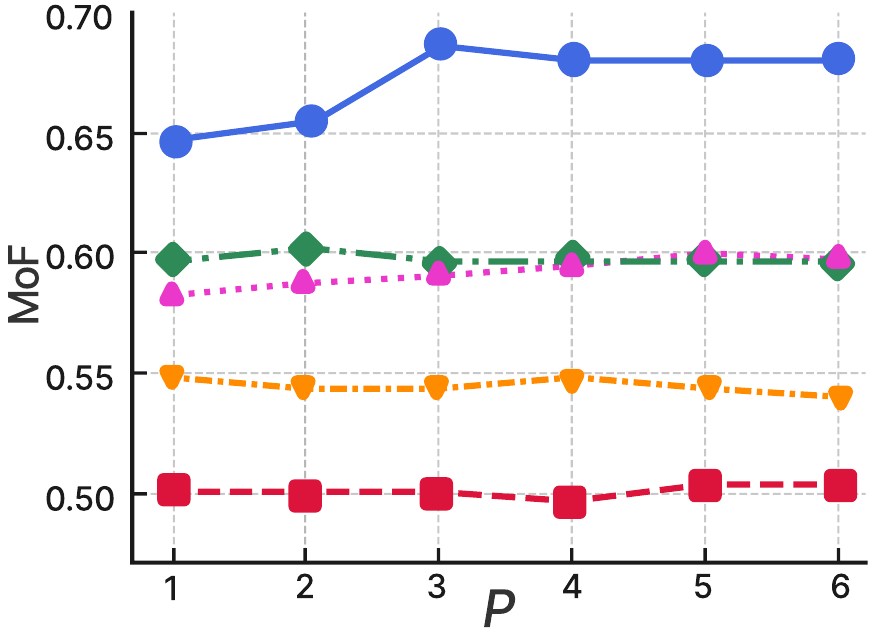}
    \end{subfigure}
    
 \vspace{-0.8em}   
 \caption{Sensitivity analysis for MoF, IoU and F1 on $\text{nseg}$ and  and MoF on $P$. \\\textcolor{RoyalBlue}{\textbf{DA} ($\circ$, solid)}, \textcolor{Red}{\textbf{FS (Mid)} ($\square$, dashed)}, \textcolor{RubineRed}{\textbf{FS (Eval)} ($\triangle$, dotted)}, \textcolor{ForestGreen}{\textbf{BF} ($\Diamond$, dash-dot)}, and \textcolor{BurntOrange}{\textbf{YTI} ($\triangledown$, custom dash)}. }
 
\label{fig:sensitive_nseg_M}
\end{figure*}

\paragraph*{Effect of  $K'$.} 
$K'$ controls the capacity of the decoder, that is the  \emph{maximum} number of predicted segments, the actual number is learned.  We set $K' = K + \text{nseg}$, where $K$ is the number of action classes and $\text{nseg} \in \mathbb{Z}$ is a dataset-specific offset. See the results for $\text{nseg} \in [-6, +6]$ in Fig.~\ref{fig:sensitive_nseg_M} (first three tables). The straight lines indicate the SOTA. CLOT achieves equal or superior performance in F1 and mIoU across nearly all $\text{nseg}$ in BF, 50S(Mid), 50S(Eval), and DA. For MoF, CLOT consistently outperforms or equals the SOTA $\forall \text{nseg}$ in BF, 50S (Mid), and YTI. 
 In particular, negative $\text{nseg}$ can help reduce over-segmentation in presence of background or very unbalanced action durations. 

 \paragraph{Effect of $M$.} We apply SWD exclusively to the initial frame-to-action matching, as this is where raw visual features are directly compared to action prototypes. At this stage, SW best captures the underlying geometric structure and distributional structure by projecting features onto multiple one-dimensional subspaces. 
Later stages use refined embeddings where this structure is no longer preserved, making SWD less meaningful.
We use $M = P \times d$ projections, where $d$ is the feature dimension and with $M \in [50, 100]$ typically sufficient for convergence (see \cite{kolouri2019generalized,deshpande2019maxsliced}). As shown in Fig.~\ref{fig:sensitive_nseg_M}, we observe stable performance across $P\in [1, 6]$.

\section{Complexity}
CLOT introduces a multi-level cyclic learning mechanism, solving three OT problems instead of one, which could lead to increased computational costs. However, to maintain efficiency, we incorporate an entropy-regularized solver, similar to ASOT~\cite{Xu2024}, which accelerates the OT optimization process while maintaining accuracy. This approach leverages sparsity structures in the cost matrices, enabling faster convergence with a computational complexity of $O(NK)$ per iteration. As a result, CLOT remains scalable to large datasets, efficiently handling videos with thousands of frames while keeping processing times competitive. The overall runtime remains comparable to ASOT, with only a moderate increase in memory usage due to additional attention layers in the decoder. Tab.~\ref{tab:flops_comparison} shows the GFLOPs and per-batch training time, using \texttt{flop counter} in PyTorch. 
Despite CLOT using 3 Optimal Transport (OT) problems and adding architectural complexity, it increases Floating Point Operations (FLOPs) and runtime moderately.

\begin{table}[h]
\centering
\resizebox{1.0\columnwidth}{!}{%
\begin{tabular}{lcccccc}
\toprule
 & Time (s) & BF & YTI & 50S (Eval) & 50S (Mid) & DA \\
\midrule
\textbf{ASOT}& 2.2 & 24.7 & 110.7 & 27.8 & 36.0 & 86.8 \\
\textbf{CLOT}& 3.7 & 48.4 & 117.7 & 36.9 & 62.8 & 125.8 \\
\bottomrule
\end{tabular}
}
\vspace{-0.9em}
\caption{GigaFLOPs and the average training runtime.}
\label{tab:flops_comparison}
\vspace{-1.5em}
\end{table}

\section{Additional Ablation Study}

\paragraph{Parallel decoding vs. autoregressive decoding}
To assess the influence of decoding strategies on segmentation quality, we compare our parallel decoder with its autoregressive counterpart, similar to~\cite{vaswani2017} across both activity-level and video-level evaluations (Tab~\ref{tab:sensitive_analysis_decoder}). The parallel design, aligned with CLOT’s architecture, avoids error propagation by predicting segments simultaneously rather than sequentially. Results indicate consistent improvements across all datasets and metrics, with notable gains in F1 and mIoU, especially for short or overlapping actions, highlighting the benefit of structured segment-level modelling over temporally fragile autoregressive predictions.
 \begin{table*}[h]
    \centering
    \setlength\tabcolsep{3pt}
    \resizebox{6.0in}{!}{
    \begin{tabular}{@{}llcccccccccccccccc@{}}
    \toprule
    \multirow{2}{*}{Eval}  &\multirow{2}{*}{Decoder} & \multicolumn{3}{c}{Breakfast} &  \multicolumn{3}{c}{YTI} &\multicolumn{3}{c}{50Salads (Mid)} & \multicolumn{3}{c}{50Salads (Eval)} & \multicolumn{3}{c}{DA} \\
    \cmidrule(lr){3-5} \cmidrule(lr){6-8} \cmidrule(lr){9-11} \cmidrule(lr){12-14} \cmidrule(lr){15-17} 
     & & MoF & F1 & mIoU & MoF & F1 & mIoU & MoF & F1 & mIoU & MoF & F1 & mIoU & MoF & F1 & mIoU  \\
    \midrule
    \multirow{2}{*}{Activity} & Autoregressive& 55.6 & 35.2 & 16.6 & 51.5 & 34.4 & 20.6 & 52.3 & 44.3 & 30.2 & 55.9 & 55.0 & 32.7 & 35.3 & 25.3 & 13.0 \\
    
    &\cellcolor{Gray}Parallel  & \cellcolor{Gray}\textbf{60.1}  &\cellcolor{Gray}\textbf{40.1} &  \cellcolor{Gray}\textbf{18.5} & \cellcolor{Gray}\textbf{54.4} &\cellcolor{Gray}\textbf{36.7} & \cellcolor{Gray}\textbf{23.4} & \cellcolor{Gray}\textbf{50.6}  & \cellcolor{Gray}\textbf{46.6} &\cellcolor{Gray} \textbf{31.4} & \cellcolor{Gray}\textbf{59.4} & \cellcolor{Gray}\textbf{63.2} & \cellcolor{Gray}\textbf{38.8} &\cellcolor{Gray} \textbf{68.8} & \cellcolor{Gray}\textbf{72.6} & \cellcolor{Gray}\textbf{48.1} \\
    \midrule
    \multirow{2}{*}{Video} & Autoregressive  & 63.1 & 53.5 & 34.9 & 68.7 & 60.5 & 44.2 & 66.7 & 60.0 & 41.2 & 61.3 & 62.4 & 36.4 & 42.6 & 30.1 & 15.8 \\
    &\cellcolor{Gray}Parallel & \cellcolor{Gray}\textbf{66.3} & \cellcolor{Gray}\textbf{55.9} & \cellcolor{Gray}\textbf{37.1}& \cellcolor{Gray}\textbf{69.3} &\cellcolor{Gray} \textbf{60.8} &\cellcolor{Gray} \textbf{48.2} & \cellcolor{Gray} \textbf{69.4} & \cellcolor{Gray}\textbf{63.8}  & \cellcolor{Gray}\textbf{45.0} & \cellcolor{Gray}\textbf{64.6} & \cellcolor{Gray}\cellcolor{Gray}\textbf{69.7} & \cellcolor{Gray}\textbf{42.5} & \cellcolor{Gray} \textbf{73.5}& \cellcolor{Gray}\textbf{75.2}  & \cellcolor{Gray}\textbf{52.4}  \\
    \bottomrule
    \end{tabular}} 
    \caption{Comparison between Autoregressive and Parallel decoders evaluated at activity and video level across the four datasets.}\label{tab:sensitive_analysis_decoder}
    \vspace{-4mm}
\end{table*}

\begin{figure}[t]
 \centering
 \includegraphics[width=\linewidth]{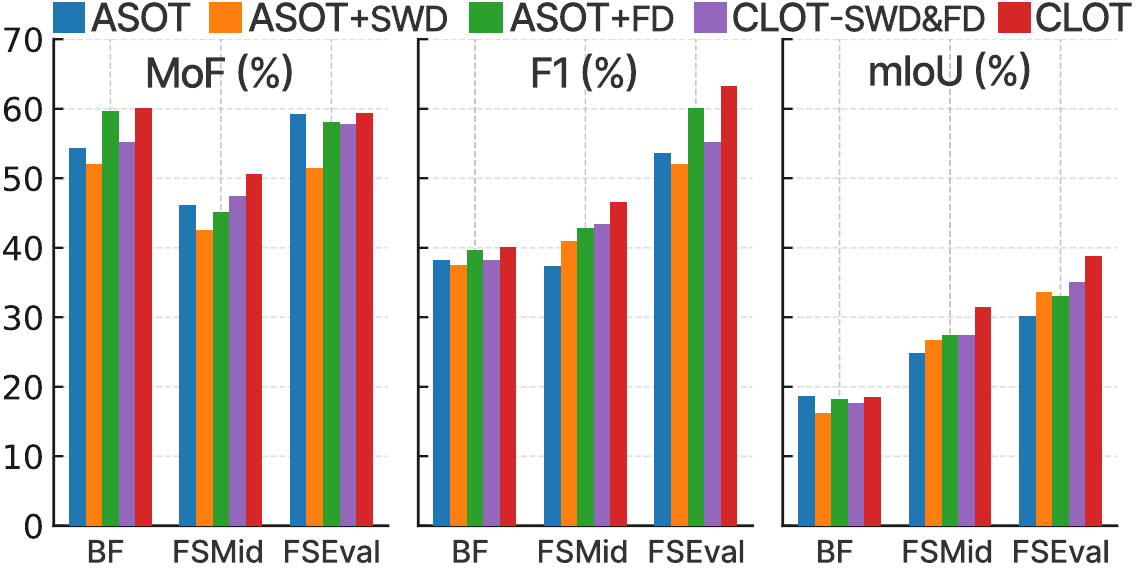}

 \vspace{-0.8em}   
 \caption{Comparative results, isolating the impact of each component.}
 
\label{fig:clotvsasot}
\end{figure}

\paragraph{Comparison to ASOT.}
ASOT corresponds to our \textcolor{gray!90}{\textbf{1st stage}} (frame-to-action) without SWD and FD. However, it is not meaningful to treat SWD or FD as independent add-ons to ASOT, as our method redefines the OT formulation itself by integrating both components to enhance representations prior to transport resolution.
These are not plug-and-play modules but part of a coherent architectural redesign. In Fig. \ref{fig:clotvsasot}, both components bring measurable benefits over ASOT, while their absence in CLOT underscores the critical role they play in overall performance. 

\section{Implementation Details}
\paragraph{Hyperparameter Settings.}
In Tab. \ref{tab: othertab}, we provide a summary of the hyperparameter settings used in our experiments. The values are categorized based on their role in the encoder, decoder, and optimal transport (OT) components of the model.

\paragraph{Computing Resources.} All experiments were conducted on a single NVIDIA GeForce RTX 3090 GPU (24GB) with CUDA 12.3, providing the necessary computational resources for training and evaluation.

\section{Additional Results}
\paragraph{Quantitative Results per Activity}
To provide a more comprehensive evaluation, we report class-wise performance for the Breakfast and YTI datasets in Tab.~\ref{tab:breakfast_yti_metrics_peract} across key metrics such as MOF, F1, and mIoU.

\paragraph{Qualitative Results}
\begin{figure*}[t]
    \centering
    \begin{subfigure}[t]{0.48\linewidth}
        \includegraphics[width=\linewidth]{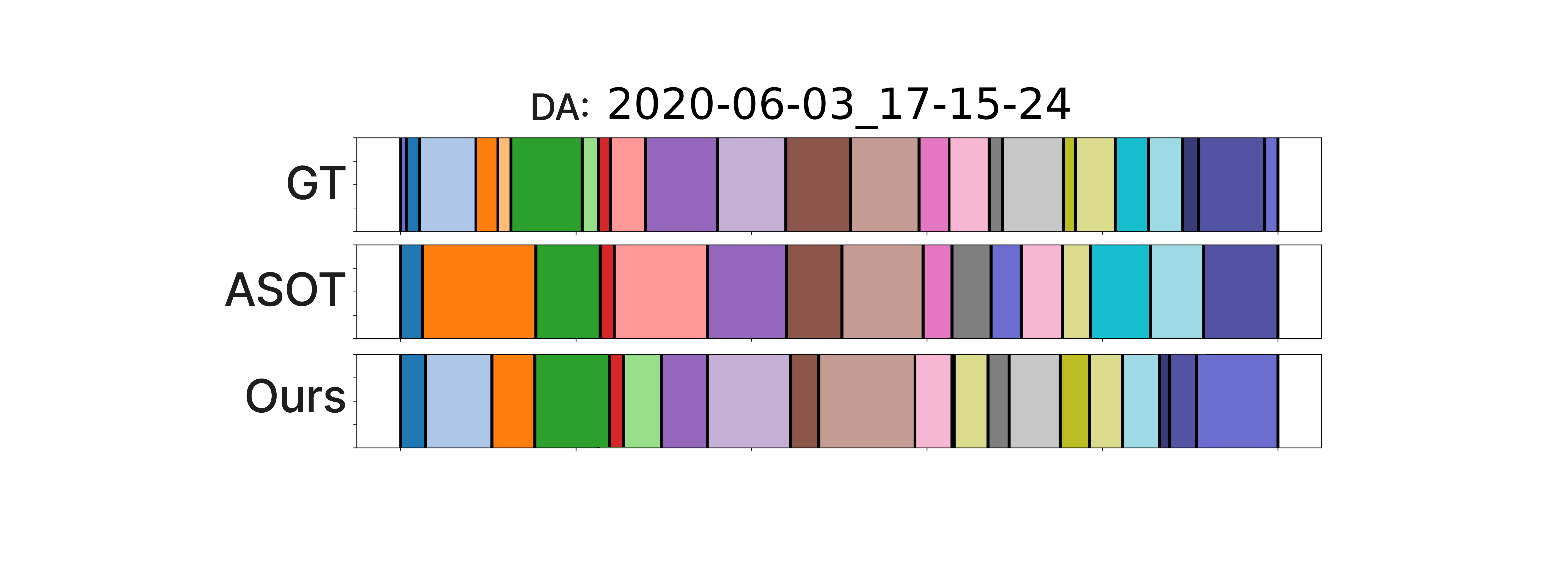}
    \end{subfigure}
    \begin{subfigure}[t]{0.48\linewidth}
        \includegraphics[width=\linewidth]{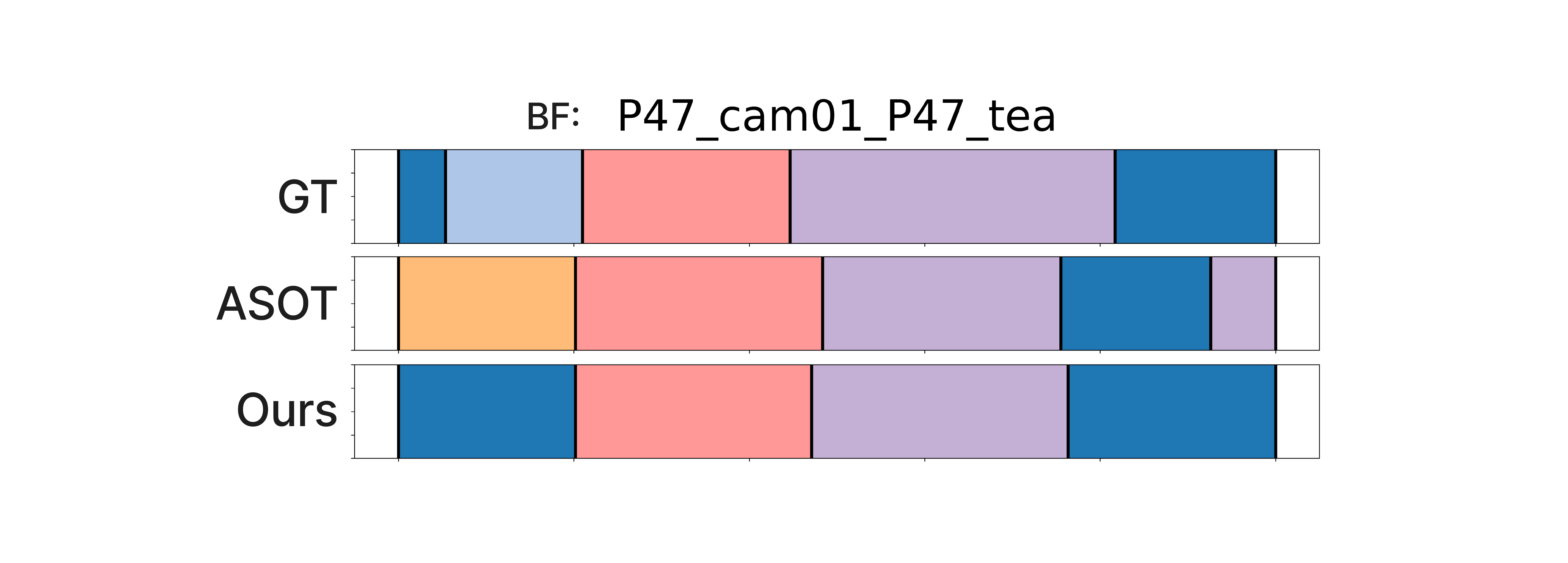}
    \end{subfigure}

    \begin{subfigure}[t]{0.48\linewidth}
        \includegraphics[width=\linewidth]{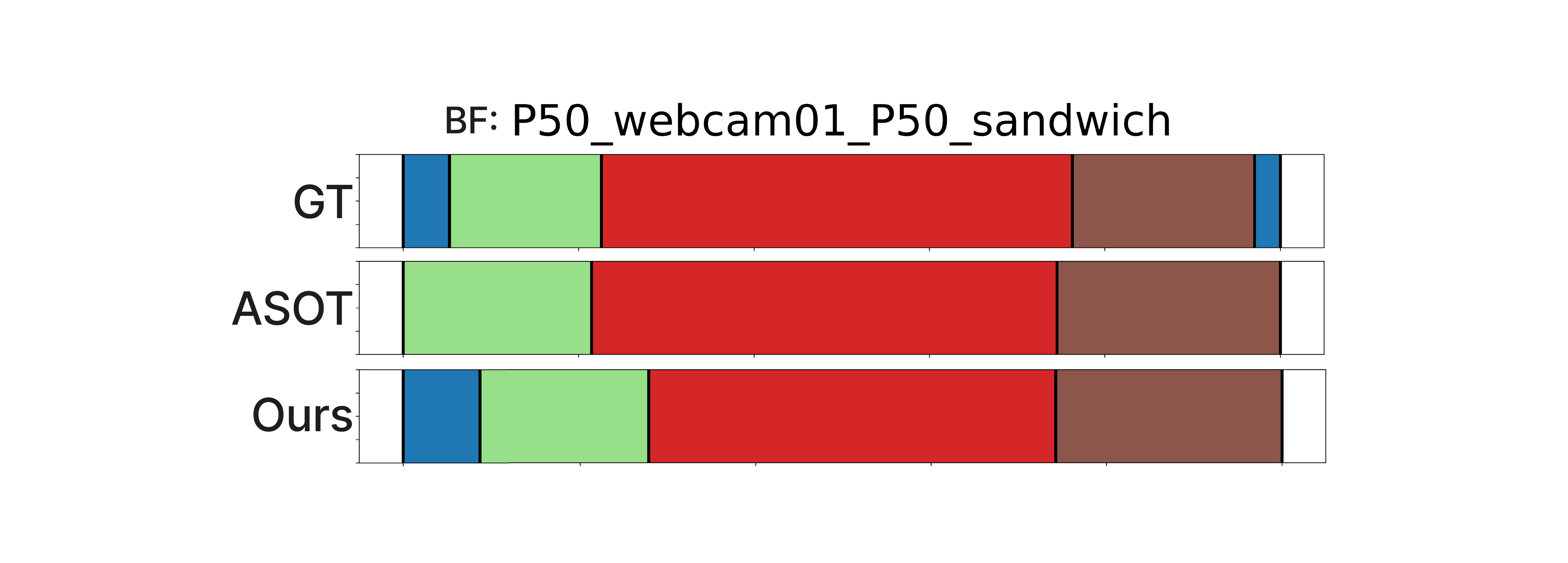}
    \end{subfigure}
    \begin{subfigure}[t]{0.48\linewidth}
        \includegraphics[width=\linewidth]{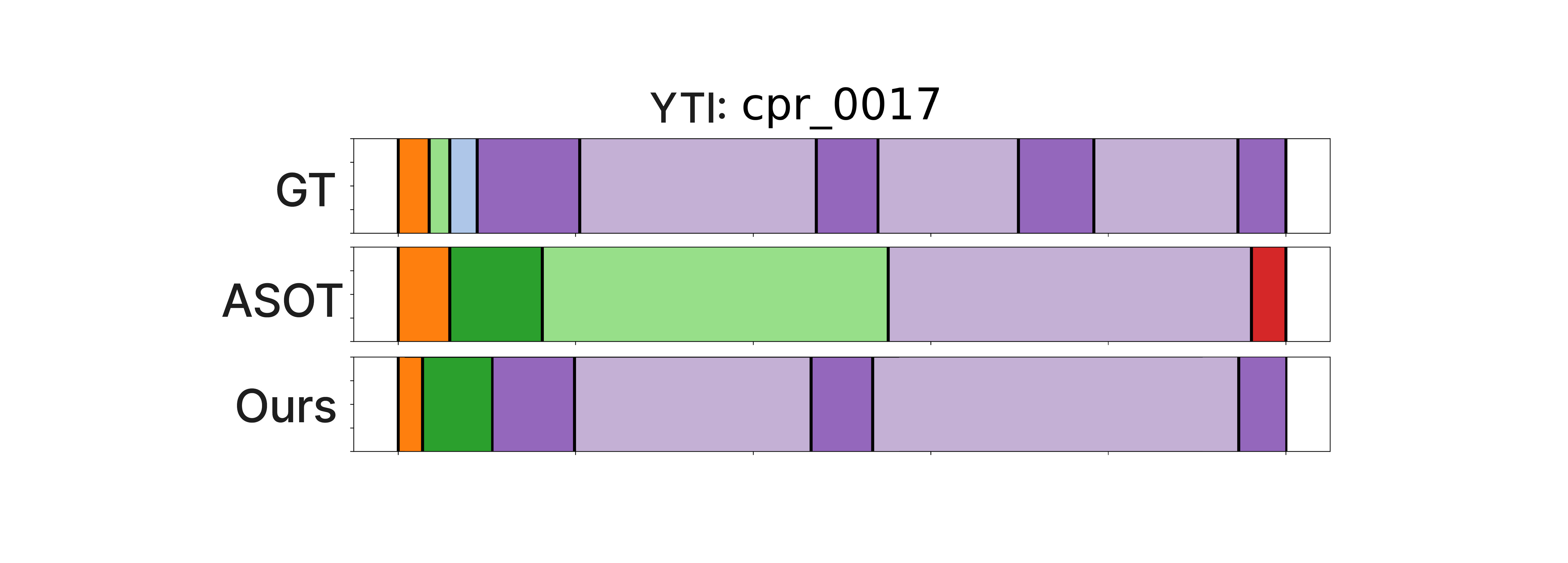}
    \end{subfigure}

    \begin{subfigure}[t]{0.48\linewidth}
        \includegraphics[width=\linewidth]{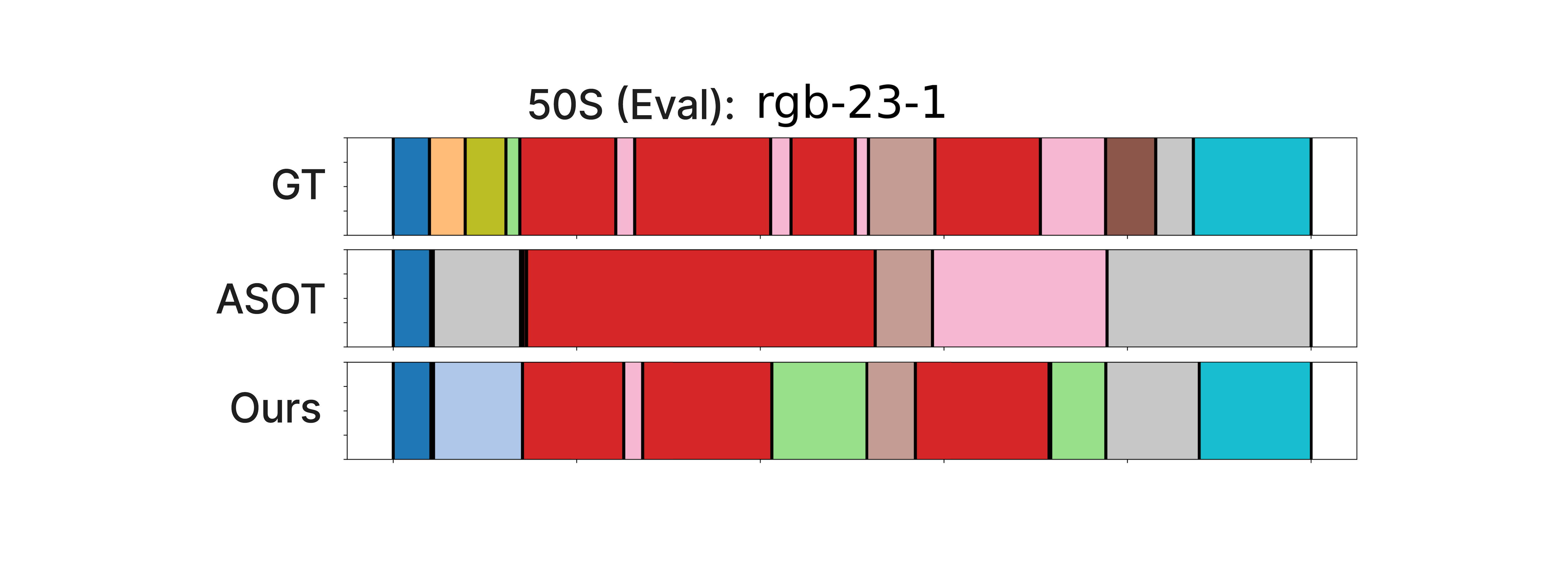}
    \end{subfigure}
    \begin{subfigure}[t]{0.48\linewidth}
        \includegraphics[width=\linewidth]{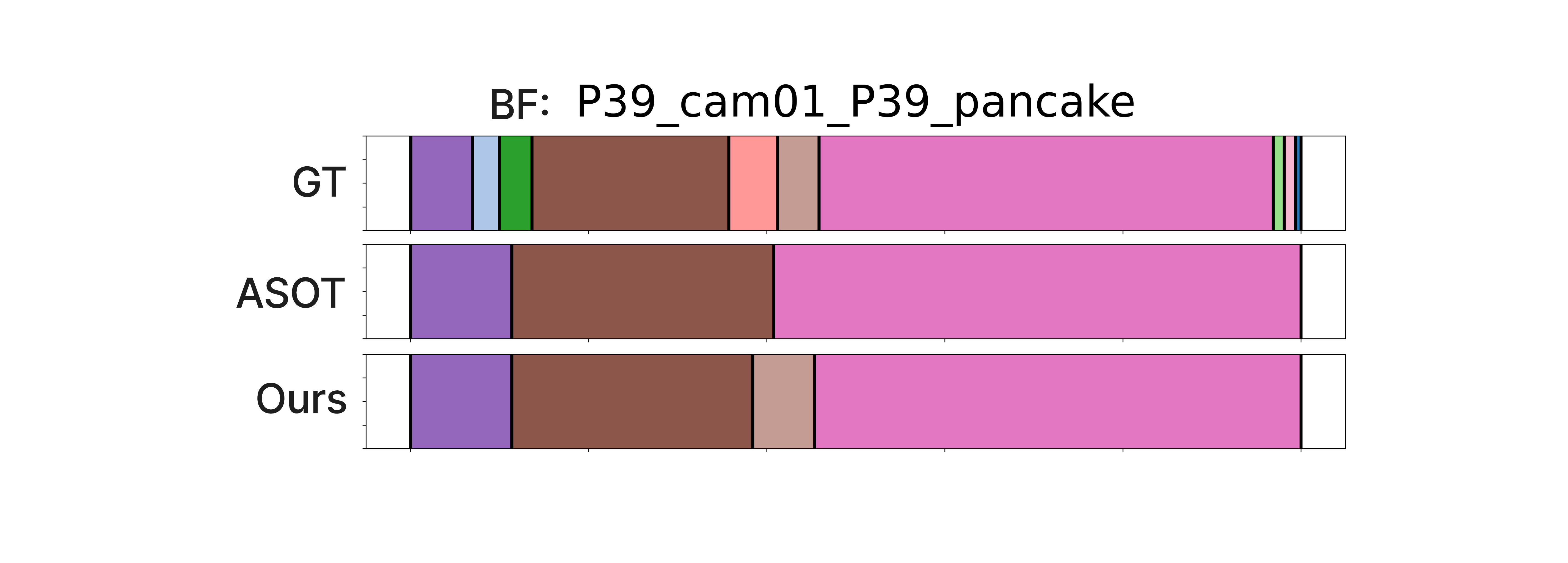}
    \end{subfigure}
\caption{\textbf{Qualitative results.} We display the ground-truth (GT), the results of CLOT (Ours), and ASOT~\cite{Xu2024}. Results from different datasets and activities for comparison.
}
\vspace{-4mm}
 \label{fig:qualitative_supp}
\end{figure*}
We also present additional qualitative examples across all four datasets in Fig.~\ref{fig:qualitative_supp}. These visualizations demonstrate CLOT’s ability to more accurately delineate action boundaries when compared to ASOT and other baseline methods, highlighting its superior segmentation performance.

\begin{table}[t]
\vspace{-12em}
\centering
\begin{tabular}{p{0.45\linewidth}ccc}
\hline
\multicolumn{4}{c}{\textbf{Breakfast Dataset}} \\
\hline
\textbf{Activity} & \textbf{MoF} & \textbf{F1} & \textbf{mIoU} \\
\hline 
coffee       & 39.4 & 26.3 & 17.0 \\
cereals      & 60.4 & 51.1 & 29.0 \\
tea          & 58.2 & 48.0 & 24.0 \\
milk         & 51.0 & 48.0 & 24.9 \\
juice        & 73.5 & 46.9 & 56.3 \\
sandwich     & 67.9 & 37.6 & 53.4 \\
scrambledegg & 51.2 & 34.1 & 52.2 \\
friedegg     & 65.1 & 39.1 & 48.2 \\
salat        & 43.7 & 41.2 & 57.2 \\
pancake      & 68.3 & 29.5 & 36.3 \\
 
\hline
\multicolumn{4}{c}{\textbf{YTI Dataset}} \\
\hline
\textbf{Activity} & \textbf{MoF} & \textbf{F1} & \textbf{mIoU} \\
\hline  
changing tire & 49.9 & 30.3 & 21.0 \\
coffee        & 67.2 & 44.3 & 30.9 \\
cpr           & 60.2 & 46.8 & 23.0 \\
jump car      & 35.7 & 26.3 & 17.0 \\
repot         & 43.7 & 27.2 & 14.0 \\
\end{tabular}
\caption{Per-class performance on the Breakfast~\cite{breakfast} and Youtube Instr. ~\cite{ytii} datasets.}
\label{tab:breakfast_yti_metrics_peract}
\end{table}
\begin{table*}[ht]
    \centering
    
    \begin{tabular}{lc}
        \hline
        \textbf{Hyperparameter} & \textbf{Value}   \\
        \hline
        Learning Rate & 0.001 (BF and DA), 0.005 (YTI and 50S) \\
        Batch Size & 2 \\
        Epochs & 15(BF), 20 (YTI), 30(50S), 70 (DA) \\
        Optimizer & Adam \\
        Loss Function & Cross-Entropy \\
        Dropout Rate & 0.5 \\
        Weight Decay & $1e^{-4}$ \\
        Activation Function & ReLU \\
        Decoder: heads & 8\\
         Decoder: dropout & 0.5(BF), 0.2(YTI), 0.1(50S-Mid), 0.2(50S-Eval), 0.5(DA)\\
        Decoder: Num of layer &2(BF), 2(YTI), 1(50S-Mid), 4(50S-Eval), 3(DA)\\
        $nseg$  & 0(BF), -6(YTI), 2(50S-Mid), -6(50S-Eval), 0(DA)\\
        $P$  & 2(BF), 2(YTI), 1(50S-Mid), 4(50S-Eval), 3(DA)\\
        $\rho$ & 01.5(BF), 0.2(YTI), 0.1(50S-Mid), 0.1(50S-Eval), 0.25(DA)\\
        $\lambda$ & 0.1(BF), 0.08(YTI), 0.11(50S-Mid), 0.2(50S-Eval), 0.16(DA)  \\
        $Nr$ & 0.04(BF), 0.02(YTI), 0.1(50S-Mid), 0.1(50S-Eval), 0.25(DA)  \\
        
        \hline
    \end{tabular}   

    \caption{Hyperparameter Settings}
    \label{tab: othertab}
\end{table*}

\clearpage
\end{document}